\documentclass[sigconf, nonacm]{acmart}

\usepackage{booktabs}   
\usepackage{subcaption} 
\usepackage{amsmath}  
\usepackage{bm}
\usepackage{algorithm}
\usepackage{algpseudocode}

\newcommand{\mymat}[1]{\mathbf{#1}}
\newcommand{\myvec}[1]{\mathbf{#1}}

\usepackage{caption}
\usepackage{listings}
\usepackage{xcolor}
\usepackage{graphicx}
\usepackage{xspace}
\usepackage{multirow}
\usepackage{centernot}
\usepackage{amsthm}
\usepackage{fontawesome}
\usepackage{url}
\usepackage{float}
\usepackage{pgfplots}
\usepackage{balance}
\pgfplotsset{compat=1.18}
\usepgfplotslibrary{groupplots}
\newcommand{\nnz}{\textit{nnz}}

\usepackage{enumitem} %for description environment
\setlist{leftmargin=*}

\newcommand{\metacomment}[3]%
{{\color{#1}{\bf #2}:\ #3}}
\newcommand{\mat}[1]{\mathbf{#1}}

\algrenewcommand\algorithmiccomment[1]{\hfill$\langle$\textsf{#1}$\rangle$}
\algrenewcommand\algorithmicrequire{\textbf{Input:}}
\algrenewcommand\algorithmicensure{\textbf{Output:}}

\algtext*{EndFunction}

\begin{document}
\title[Scalable Optimal Transport Algorithm for Network Alignment]{Scalable Optimal Transport Algorithm for Network Alignment}

\author{Elaheh Hassani$^{*}$}
\affiliation{
    \institution{Texas A\&M University}
    \state{TX}
    \country{USA}}
\email{ehassani@tamu.edu}

\author{Durga Mandarapu$^{*}$}
\affiliation{
    \institution{Lawrence Berkeley National Laboratory}
    \state{CA}
    \country{USA}}
\email{durga@lbl.gov}

\author{Qi Yu}
\affiliation{
    \institution{University of Illinois at Urbana-Champaign}
    \state{IL}
    \country{USA}}
\email{qiyu6@illinois.edu}

\author{Hanghang Tong}
\affiliation{
    \institution{University of Illinois at Urbana-Champaign}
    \state{IL}
    \country{USA}}
\email{htong@illinois.edu}

\author{Ariful Azad}
\affiliation{
    \institution{Texas A\&M University}
    \state{TX}
    \country{USA}}
\email{ariful@tamu.edu}
\begin{abstract}
    Network alignment identifies node correspondences across different networks and is a fundamental primitive in many data science applications, including social network analysis, fraud detection, and knowledge graph integration. 
    However, state-of-the-art network alignment methods often achieve high accuracy by repeatedly constructing and updating dense matrices, sacrificing scalability in the process.
    To address this scalability limitation without compromising alignment accuracy, we present \textsc{FastAlign}, a scalable, sparsity-aware framework for optimal transport-based network alignment.
    Rather than introducing a new alignment model, \textsc{FastAlign} preserves the original OT formulation and reinterprets its computation as a set of recurring mixed sparse-dense operations. 
    \textsc{FastAlign} combines sparsity-aware graph computation with domain-specific kernel fusion, including a custom SpMM kernel.
    Our results show that \textsc{FastAlign} achieves alignment quality comparable to state-of-the-art OT-based methods while substantially reducing end-to-end runtime up to $3.89\times$--$9.45\times$ on CPU and $2.24\times$--$32.54\times$ on GPU. 
\end{abstract}

\maketitle

\renewcommand\thefootnote{*}\footnote{These authors contributed equally.}\addtocounter{footnote}{-1}

\noindent\textbf{Code Availability:} The source code, data, and other artifacts 
are available at \url{https://github.com/elawh1/FastAlign}.

\section{Introduction}

The goal of network alignment is to identify node correspondences between two networks. 
It is a fundamental primitive for integrating graph-structured data across sources in many applications, including social network analysis~\cite{cao2016socialnetworkanalysis}, fraud detection~\cite{du2021frauddetection}, and knowledge graphs~\cite{wang2018knowledgegraph}. 
For example, aligning users across social platforms supports cross-platform recommendations; matching entities across transaction networks helps detect suspicious activities; aligning entities across incomplete knowledge graphs enables the construction of unified knowledge bases.
As modern networks grow in size and dynamic networks require repeated realignments, it is imperative that alignment algorithms be both accurate and fast.
% Scalability is often a central requirement in network alignment applications as the input networks are often large and sparse. 
% In these applications, alignment often is not a one-time offline task and may require repeated realignment as the graphs grow.
% Therefore, at modern data scales, network alignment algorithms must be accurate, fast, and scalable. 

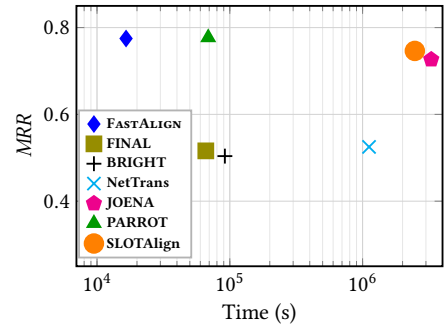
\begin{figure}[t]
\centering
\scalebox{0.7}{
\begin{tikzpicture}
\begin{axis}[
    width=8.5cm, height=6.5cm,
    xlabel={Time (s)},
    ylabel={MRR},
    label style={font=\fontsize{12}{14}\selectfont},
    ylabel style={font=\itshape\fontsize{12}{14}\selectfont},
    tick label style={font=\itshape\fontsize{12}{14}\selectfont},
    xmode=log,                    
    log basis x=10,
    xmin=7e3, xmax=4e6,             
    ymin=0.25, ymax=0.85,            
    enlargelimits=false,
    axis on top,                    
    clip=true,                      
    grid=both,
    grid style={line width=.1pt, draw=gray!20},
    major grid style={draw=gray!35},
    tick align=inside,
    legend pos=south west,        
    legend cell align={left},
    legend style={font=\bfseries\small, draw=gray!60, xshift=-5pt,},
    only marks,
    mark size=4pt,
    line width=1pt,
]
% --- one \addplot per model: coordinate is (runtime_seconds, MRR) ---
% GGI
% \addplot[mark=diamond*, color=blue]  coordinates {(21125, 0.887788)}; \addlegendentry{Ours}
% \addplot[mark=*,        color=black]  coordinates {(137312, 0.454828)}; \addlegendentry{BRIGHT}
% \addplot[mark=square*,  color=olive] coordinates {(90062, 0.338331)}; \addlegendentry{FINAL}
% \addplot[mark=pentagon*,color=magenta] coordinates {(3627452, 0.37103)}; \addlegendentry{JOENA}
% \addplot[mark=x, mark size=5pt, color=cyan] coordinates {(2510953, 0.499688)}; \addlegendentry{NetTrans}
% \addplot[mark=triangle*,color=green!60!black] coordinates {(87975, 0.854821)}; \addlegendentry{PARROT}
% \addplot[mark=+, mark size=5pt, color=orange] coordinates {(3046997, 0.740144)}; \addlegendentry{SLOTAlign}
% ACM
\addplot[mark=diamond*, color=blue]  coordinates {(16550, 0.774822)}; \addlegendentry{\textsc{FastAlign}}
\addplot[mark=square*,  color=olive] coordinates {(65986, 0.515928)}; \addlegendentry{FINAL}
\addplot[mark=+,        color=black]  coordinates {(91842, 0.503853)}; \addlegendentry{BRIGHT}
\addplot[mark=x, mark size=5pt, color=cyan] coordinates {(1118870, 0.525013)}; \addlegendentry{NetTrans}
\addplot[mark=pentagon*,color=magenta] coordinates {(3289568, 0.726521)}; \addlegendentry{JOENA}
\addplot[mark=triangle*,color=green!60!black] coordinates {(68799, 0.777059)}; \addlegendentry{PARROT}
\addplot[mark=*, mark size=5pt, color=orange] coordinates {(2474697, 0.746251)}; \addlegendentry{SLOTAlign}
\end{axis}
\end{tikzpicture}
}
\caption{MRR (accuracy) vs. runtime across network alignment methods on the ACM-DBLP networks. FINAL is consistency-based, BRIGHT and NetTrans are embedding-based, while the rest are OT-based methods.}
\label{fig:mrr-vs-runtime}
\end{figure}

Existing network alignment methods exhibit a tradeoff between scalability and accuracy (Figure~\ref{fig:mrr-vs-runtime}). 
Consistency-based methods use a fast linear transformation between two networks, although their accuracy suffers because they do not fully capture global structure~\cite{zhang2016final}. 
Embedding-based methods improve accuracy by learning node representations, at the cost of higher runtime from deep representation learning~\cite{yan2021bright, zhang2020nettrans, zhang2021nextalign}. 
Optimal transport (OT)-based methods achieve state-of-the-art (SOTA) empirical accuracy by optimizing a transport objective that combines node similarity, structural consistency, neighborhood consistency, and anchor supervision~\cite{zeng2023parrot, yu2025joena, zeng2024hot, tang2023slotalign}. 
This accuracy comes with its own scalability limitation as OT-based methods repeatedly construct and update dense cross-network cost and alignment matrices over several iterations, making them expensive in both runtime and memory.
% For large graphs, these dense iterative computations in OT-based methods create substantial runtime overhead limiting their scalability and practical use.

To address this scalability bottleneck without sacrificing alignment accuracy, we present \textsc{FastAlign}, a scalable sparsity-aware framework for OT-based network alignment. 
Rather than introducing a new alignment model, \textsc{FastAlign} preserves the original OT formulation and reinterprets its computation as a set of recurring mixed sparse-dense operations. 
This reinterpretation exposes where graph sparsity can be exploited and where dense cross-network operations must be optimized for memory efficiency. 
\textsc{FastAlign} realizes this decomposition using sparsity-aware graph operations and memory-efficient kernel fusion to reduce memory traffic and avoid unnecessary intermediate materialization. 
%As a result, \textsc{FastAlign} emerges as one of the fastest and most accurate network alignment algorithms, as shown in Figure~\ref{fig:mrr-vs-runtime}.
As a result, \textsc{FastAlign} accelerates and scales OT-based network alignment, making it practical at large graph sizes.

Building on this reinterpretation, \textsc{FastAlign} combines four optimizations. 
First, it uses sparsity-aware computations to replace all dense matrix computations involving graph structures.
%over nonexistent edges, as the real world graphs in network alignment are often 
% less than $1\%$
%more than $99\%$ sparse. 
Second, it introduces a custom sparse-dense matrix multiplication (SpMM) algorithm for multiplying a sparse matrix by a wide dense matrix, as arises in OT-based alignment, unlike generic SpMM settings where the dense matrix is tall and skinny.
Third, it uses domain-specific kernel fusion to combine chains of  
% dense element-wise operations and reductions in the cost-update and Sinkhorn stages, 
memory-bound operations to reduce intermediate materialization and memory traffic. 
 % reducing redundant memory traffic and per-operation overhead.
Fourth, its GPU implementation keeps data resident on device and reuses dense matrices across stages to avoid unnecessary data movement and tiles kernels for optimized performance.

We implement \textsc{FastAlign} on CPU and GPU and evaluate it on a range of real-world and synthetic network alignment datasets. 
Our results show that \textsc{FastAlign} achieves comparable alignment quality to SOTA OT-based methods while substantially reducing end-to-end runtime. 
% In particular, \textsc{FastAlign} achieves $3.89\times$--$9.45\times$ speedup on CPU and $32.5\times$ speedup on GPU, demonstrating that high-accuracy OT-based network alignment can be made significantly more practical through sparsity-aware and memory-efficient reinterpretation.

In summary, this paper makes the following contributions:

\begin{itemize}

\item We present \textsc{FastAlign}, a scalable sparsity-aware framework for OT-based network alignment that preserves the alignment accuracy while improving runtime and memory efficiency.

\item We provide a computational reinterpretation of OT-based network alignment using mixed sparse-dense operations. 

\item We develop sparsity-aware graph computation, including a custom SpMM kernel and the fusion of adjacent memory-bound operations, to accelerate OT-based network alignment. 

\item With extensive experiments, we demonstrate that \textsc{FastAlign} achieves $3.89\times$--$9.45\times$ speedup on CPU and $2.24\times$--$32.54\times$ speedup on GPU relative to SOTA network alignment algorithms. 

% \item We provide a computational reinterpretation of OT-based network alignment, showing that its execution is dominated by a small set of recurring mixed sparse-dense operations.

% \item We introduce sparsity-aware graph computation into OT pipeline to avoid dense matrix computations over sparse input graphs.

% \item We develop a custom SpMM kernel for sparse graph matrices multiplied with wide dense cross-network matrices, making the SpMM more cache friendly on CPU.

% \item We design fused kernels for memory-bound dense operations in cost computation and Sinkhorn optimization, reducing intermediate materialization and repeated memory accesses of matrices.

\end{itemize}

% \vspace{-3pt}
\section{Background}
\label{sec:background}
% \metacomment{blue}{todo}{one sentence about network alignment applications and then say we are defining problems.}
% Network alignment supports many multi-network mining tasks, such as linking users across social platforms and matching proteins across biological networks.
% In this section, we provide a brief background with definitions of network alignment and optimal transport formulations.

\begin{figure}
  \centering
  \includegraphics[width=0.85\linewidth]{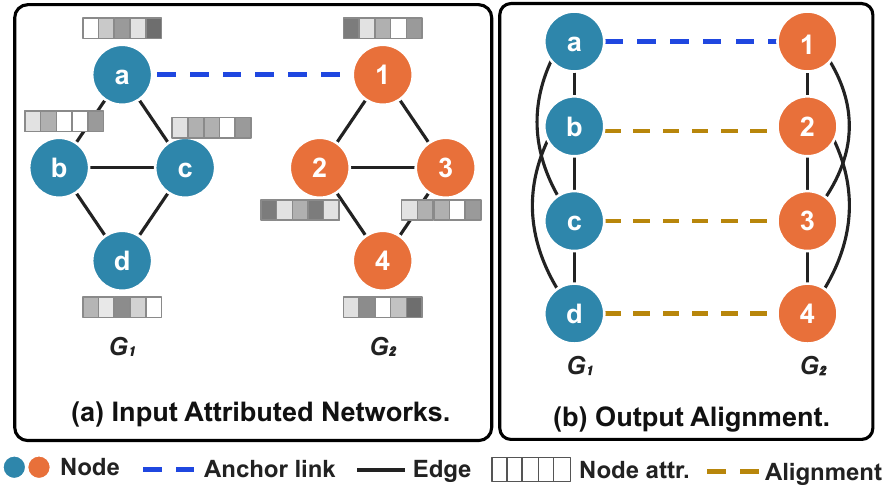}
  \caption{Network alignment problem. Given two attributed networks $\mymat{G}_1$ and $\mymat{G}_2$ and a set of anchor links, network alignment seeks a correspondence of nodes across networks.}
  \label{fig1:na}
\end{figure}

%\paragraph{Network alignment problem.}
Let $\mymat{G}_1 = (\mymat{V}_1, \mymat{E}_1, \mymat{A}_1)$ and
$\mymat{G}_2 = (\mymat{V}_2, \mymat{E}_2, \mymat{A}_2)$ denote two networks, with $|\mymat{V}_1| = n_1$, $|\mymat{V}_2| = n_2$,
and adjacency matrices $\mymat{A}_1 \in \mathbb{R}^{n_1 \times n_1}$ and
$\mymat{A}_2 \in \mathbb{R}^{n_2 \times n_2}$.
Nodes optionally
carry attributes encoded as 
$\mymat{X}_1 {\in} \mathbb{R}^{n_1 \times d}$ and
$\mymat{X}_2 \in \mathbb{R}^{n_2 \times d}$.
We are further given a set of anchor pairs $\mathcal{L} \subseteq \mymat{V}_1 \times \mymat{V}_2$, where each pair denotes a known correspondence between the two networks.
Network alignment computes a soft alignment
matrix $\mymat{S} \in \mathbb{R}^{n_1 \times n_2}$, where
$\mymat{S}_{ij}$ scores likelihood of node $i \in \mymat{V}_1$
corresponding to node $j \in \mymat{V}_2$.
Figure~\ref{fig1:na} illustrates network alignment problem. 
Existing approaches can be categorized as consistency-based, embedding-based, or OT-based~\cite{zhang2020network}. 
Of these, OT-based approaches consistently achieve SOTA performance~\cite{yu2025planetalign}, our algorithm builds on this paradigm. 
So, we detail  OT-based methods in this section and defer the rest to Section~\ref{sec:related_work}.

\vspace{-4pt}
\subsection{Optimal transport for network alignment}
% Optimal transport (OT), also known as the Wasserstein distance, maps
% one discrete distribution to another by minimizing a transport
% cost~\cite{peyre2019computational}.
Optimal transport (OT) maps one discrete distribution to another by minimizing a total transport
cost, known as the Wasserstein distance~\cite{peyre2019computational}.
% \qi{I think it's better to describe OT in this way: Optimal transport (OT) maps one discrete distribution to another by minimizing a total transport cost, known as the Wasserstein distance}.
OT-based network alignment treats
the node sets $\mymat{V}_1$ and $\mymat{V}_2$ as discrete
distributions with uniform mass vectors
$\boldsymbol{\mu} = \tfrac{1}{n_1}\mymat{1}_{n_1}$ and
$\boldsymbol{\nu} = \tfrac{1}{n_2}\mymat{1}_{n_2}$. The soft alignment
matrix $\mymat{S}$ corresponds to the OT transport plan, whose entry
$\mymat{S}_{ij}$ encodes the amount of mass routed from node
$i \in \mymat{V}_1$ to node $j \in \mymat{V}_2$ and is interpreted as
the strength of their correspondence. Given a cost matrix
$\mymat{C} \in \mathbb{R}^{n_1 \times n_2}$ that encodes the penalty
for aligning each pair of nodes, the optimal alignment is obtained by
solving
\begin{equation}
  \min_{\mymat{S} \in \Pi(\boldsymbol{\mu}, \boldsymbol{\nu})}
  \; \langle \mymat{S}, \mymat{C} \rangle - \varepsilon\, H(\mymat{S}),
  \label{eq:entropic-ot}
\end{equation}
where $\langle \cdot \rangle$ denotes the Frobenius inner product, $\Pi(\boldsymbol{\mu}, \boldsymbol{\nu}) = \{ \mymat{S} \in
\mathbb{R}_{\geq 0}^{n_1 \times n_2} : \mymat{S}\mymat{1} =
\boldsymbol{\mu},\; \mymat{S}^\top \mymat{1} = \boldsymbol{\nu} \}$
is the set of feasible soft alignments, and $\varepsilon > 0$
controls the strength of the entropic regularizer $H(\mymat{S})$ that
makes the problem strictly convex and efficiently solvable by the
Sinkhorn algorithm~\cite{cuturi2013sinkhorn}.
The objective in Eq. ~\ref{eq:entropic-ot} is shaped by two design choices: how the cost matrix $\mymat{C}$ is constructed and how the resulting optimization is solved. 
We discuss each in turn.

% \metacomment{blue}{todo}{transition sentences}

\subsection{Decomposition of the Cost Matrix}
\label{sec:ot-cost}

% The effectiveness of OT-based network alignment hinges on the design of the cost matrix $\mymat{C}$. 
The design of the cost matrix $\mymat{C}$ is what most directly shapes alignment quality.
A well-designed cost captures three consistency principles together with prior anchor knowledge ~\cite{zeng2023parrot}.
\emph{Node consistency} ($\mymat{C}_{\text{node}}$) requires matched
nodes to have similar attributes and similar structural positions
within their respective networks. \emph{Edge consistency}
($\mymat{C}_{\text{edge}}$) requires matched node pairs to preserve
their connecting edges. \emph{Neighborhood consistency}
($\mymat{C}_{\text{nbr}}$) requires matched nodes to have matched
neighborhoods, so that neighbors in one graph map to neighbors in the
other. Finally, the cost should reflect prior anchor preferences
($\mymat{C}_{\text{anc}}$), biasing the alignment toward known
correspondences~\cite{zeng2023parrot}. The overall cost matrix
combines these terms additively,
where $\lambda_e, \lambda_n, \lambda_a \geq 0$ weigh the contribution of each term:
% \begin{equation}
%   % \min_{\mymat{S}\in\Pi(\mymat{a},\mymat{b})}\;
%   % \underbrace{\langle \mymat{C}_{\text{cross}}, \mymat{S}\rangle}_{\text{node}}
%   \mymat{Cost} = \underbrace{\mymat{C}_{node}}_{node}
%   \;+\; \underbrace{\lambda_e \mymat{C}_{edge}}_{edge}
%   \;+\; \underbrace{\lambda_n\ \mymat{C}_{nbr}}_{neigborhood}
%   \;+\; \lambda_a  \underbrace{\mymat{C}_{ancr}}_{anchor},
%   \label{eq:reg-ot-na}
% \end{equation} 
\begin{equation}
  \mymat{C} =
  \mymat{C}_{\text{node}}
  \;+\; \lambda_e\, \mymat{C}_{\text{edge}}
  \;+\; \lambda_n\, \mymat{C}_{\text{nbr}}
  \;+\; \lambda_a\, \mymat{C}_{\text{anc}}
  \label{eq:reg-ot-na}
\end{equation}

% With the cost matrix $\mymat{C}$ in place, what remains is to solve Equation 1 for the alignment S.

% \metacomment{blue}{todo}{transition sentences - we dont need transition. it is taken care in start of next subsection}

\subsection{The OT solver}\label{sec:ot-solver}
Given the current cost matrix $\mymat{C}$ and the predefined marginal
vectors $\boldsymbol{\mu}, \boldsymbol{\nu}$, the Sinkhorn
algorithm computes the alignment matrix
$\mymat{S}$ by iteratively rescaling its rows and columns to satisfy
the marginal constraints while keeping the alignment cost low. 
The algorithm maintains two dual vectors $\myvec{a}$ and
$\myvec{b}$ and alternately updates them in the log domain (for numerical stability):

% \begin{equation}
% \begin{aligned}
% a_j &= \varepsilon \log \nu_j - \varepsilon \operatorname*{LSE}_{i}\!\left( \frac{b_i - \mat{C}_{ij}}{\varepsilon} \right), \\
% b_i &= \varepsilon \log \mu_i - \varepsilon \operatorname*{LSE}_{j}\!\left( \frac{a_j - \mat{C}_{ij}}{\varepsilon} \right),
% \end{aligned}
% \label{eq:sinkhorn-updates}
% \end{equation}
\begin{equation}
a_j = \varepsilon \log \nu_j - \varepsilon \operatorname*{LSE}_{i}\!\left( \tfrac{b_i - \mat{C}_{ij}}{\varepsilon} \right), \quad
b_i = \varepsilon \log \mu_i - \varepsilon \operatorname*{LSE}_{j}\!\left( \tfrac{a_j - \mat{C}_{ij}}{\varepsilon} \right)
\label{eq:sinkhorn-updates}
\end{equation}

% $a_i = \varepsilon \log \mu_i - \varepsilon \operatorname*{LSE}_{j}\!\left( \frac{b_j - \mat{Cost}_{ij}}{\varepsilon} \right)$ and $b_j = \varepsilon \log \nu_j - \varepsilon \operatorname*{LSE}_{i}\!\left( \frac{a_i - \mat{Cost}_{ij}}{\varepsilon} \right)$
where LSE is the log-sum-exp reduction. 
%The log domain keeps the computation stable when $\varepsilon$ is small.
On convergence, the alignment matrix is recovered as
% \begin{equation}
% \mymat{S}_{ij}
% = \exp\!\left( \frac{a_i + b_j - \mymat{C}_{ij}}{\varepsilon} \right).
% \label{eq:sinkhorn-plan}
% \end{equation}
\begin{equation}
\mymat{S}_{ij}
= \exp\!\left( \frac{a_j + b_i - \mymat{C}_{ij}}{\varepsilon} \right).
\label{eq:sinkhorn-plan}
\end{equation}
Eq.~\ref{eq:sinkhorn-plan} yields the optimal alignment for the
current $\mymat{C}$. Because some components of $\mymat{C}$ themselves
depend on $\mymat{S}$, the cost is then updated using the newly
computed alignment and the Sinkhorn solve is repeated. This alternation realizes the constrained proximal
point method~\cite{xie2020fast}, which solves the problem as a
sequence of fixed-cost OT subproblems with guaranteed convergence.

% Since component of cost depends on the current alignment, we alternate between recomputing the $\mymat{S}$-dependent cost terms and resolving for S with Sinkhorn until it stabilizes. 
% This alternation follows the constrained proximal point method\cite{zeng2023parrot}, which solves such $\mymat{S}$-dependent problems as a sequence of fixed-cost OT subproblems with guaranteed convergence.

\section{Scaling OT-Based Network Alignment}
\label{sec:design}
In this section, we introduce \textsc{FastAlign}, a scalable and fast network alignment framework. \textsc{FastAlign} achieves its speedups through a deliberate two-stage pipeline. 
First, it recasts OT-based alignment as a sequence of linear-algebraic operations on matrices with explicit shapes and sparsity. 
Second, it accelerates these kernels using custom sparse algorithms, kernel fusion, and hardware-mapped parallel implementations across CPUs and GPUs. 
This two-stage framework is necessary because the initial linear-algebraic decomposition exposes optimization pathways that are usually hidden by conventional OT formulations described in Section~\ref{sec:background}.

\subsection{Foundation of \textsc{FastAlign}: Linear-algebraic Decomposition of Network Alignment} \label{sec:foundation}
\begin{figure*}[!t]
  \centering
  \includegraphics[width=0.9\textwidth]{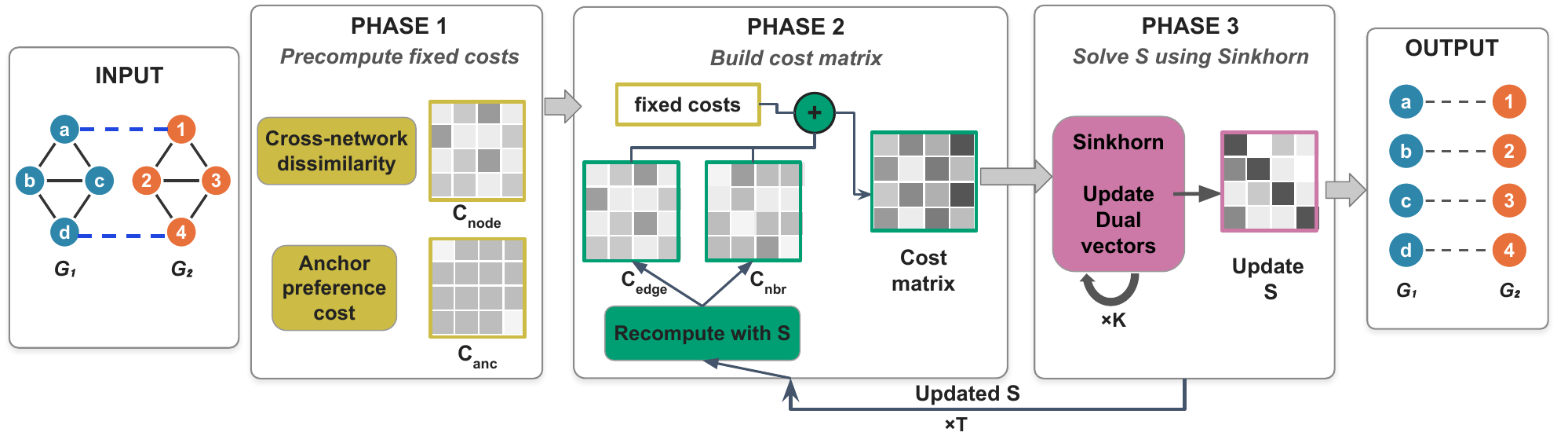}
  \caption{Overview of OT-based network alignment.
  Given two attributed networks $\mymat{G}_1$ and $\mymat{G}_2$ and two anchor links (dashed lines).
  \textbf{Phase~1} precomputes the fixed cost components, $\mymat{C}_{node}$ and $\mymat{C}_{anc}$.
  %$\mymat{C}_{node}$, capturing cross-network dissimilarity from node attributes and positional encodings, and $\mymat{C}_{anc}$, encoding prior knowledge from known anchor links.
  \textbf{Phase~2} computes two $\mymat{S}$-dependent cost terms: $\mymat{C}_{edge}$ and $\mymat{C}_{nbr}$, from the current alignment matrix $\mymat{S}$, then adds them to the fixed costs to form the Cost matrix. 
  \textbf{Phase~3} runs the Sinkhorn algorithm for $K$ iterations to optimize the dual vectors, which then update $\mymat{S}$.
  The updated $\mymat{S}$ feeds back into the $\mymat{S}$-dependent cost terms, and Phase~2-3 repeats up to $T$ outer iterations until $\mymat{S}$ converges into the final node alignment.
  }
  \label{fig:pipeline}
\end{figure*}

To decompose the OT-based algorithm in linear-algebraic form, we
first observe that the computations described in
Section~\ref{sec:background} can be organized in three phases:
(a)~Phase~1 normalizes the adjacency matrices and computes the
alignment-independent cost components $\mymat{C}_{\text{node}}$ and
$\mymat{C}_{\text{anc}}$; 
(b)~Phase~2 computes the
alignment-dependent components $\mymat{C}_{\text{edge}}$ and
$\mymat{C}_{\text{nbr}}$; and (c)~Phase~3 invokes an OT solver to
produce the alignment from the current cost matrix. 
These three phases of \textsc{FastAlign} are described in Algorithm~\ref{alg:ot-align} and Figure ~\ref{fig:pipeline}.

% A typical OT-based alignment algorithm uses a variant of the cost matrix, solved by Sinkhorn as presented in Sections~\ref{sec:ot-cost} and~\ref{sec:ot-solver}.
% This framing dominates the literature because it is natural for reasoning about correctness and convergence. 
% The mathematical formulation of Section ~\ref{sec:background} is natural for reasoning about correctness and convergence.
% However, it is not suitable for algorithmic optimization because it hides where the algorithm spends its time, and obscures the linear-algebraic kernels on which modern processors are optimized to operate.

% To make acceleration possible, we restate the algorithm as an explicit sequence of matrix operations. 
% Algorithm~\ref{alg:ot-align} presents this linear-algebraic decomposition, which is the foundation of FastAlign.
% Every sparsity, fusion, and parallelization strategy introduced in the rest of the paper is defined against this algorithm rather than against the mathematical formulation described in the background.

\algblockdefx[RepeatTimes]{RepeatTimes}{EndRepeat}[1]{\textbf{Repeat} #1 \textbf{times}:}{}
\algtext*{EndRepeat}

\begin{algorithm}[h]
\caption{FastAlign: A Linear Algebraic View of OT-Based Network Alignment}
\label{alg:ot-align}
\begin{algorithmic}[1]
\Require Graphs $\mymat{G}_1, \mymat{G}_2$; attributes
  $\mymat{X}_1, \mymat{X}_2$; anchors $\mathcal{L}$
\Ensure Alignment matrix $\mymat{S}$
\State $\bar{\mymat{A}}_1 \gets $ \Call{Rownormalized}{$\mymat{G}_1$}, \ $\bar{\mymat{A}}_2 \gets $ \Call{Rownormalized}{$\mymat{G}_2$} \label{algo-step:norm}
% \State $\mymat{R}_1\gets \Call{OneHot}{\mathcal{L}[0]}$, $\mymat{R}_2 \gets \Call{OneHot}{\mathcal{L}[1]}$ 
\State $\mymat{R}_1^{(0)}\gets \Call{OneHot}{\mathcal{L}[0]}$, $\mymat{R}_2^{(0)} \gets \Call{OneHot}{\mathcal{L}[1]}$ 
\Statex \textit{// Phase 1: precompute $\mymat{S}$-independent quantities}
% \State Perform $k$ iterations: $\mymat{R}_1\gets \bar{\mymat{A}}_1\mymat{R}_1$,\ \ $\mymat{R}_2\gets \bar{\mymat{A}}_2\mymat{R}_2$ \label{algo-step:Rk}
% \State Perform $k$ iterations: $\mathbf{R}_1 \gets (1-\beta)\,\bar{\mathbf{A}}_1 \mathbf{R}_1 + \beta\,\mathbf{R}_1^{(0)},\;\; \mathbf{R}_2 \gets (1-\beta)\,\bar{\mathbf{A}}_2 \mathbf{R}_2 + \beta\,\mathbf{R}_2^{(0)}$ \label{algo-step:Rk}
\State Perform $k$ iterations: $\mathbf{R}_i \gets (1-\beta)\,\bar{\mathbf{A}}_i \mathbf{R}_i + \beta\,\mathbf{R}_i^{(0)}$ \textbf{for} $i \in \{1,2\}$ \label{algo-step:Rk}
\State $\mymat{C}_{\text{node}} \gets
  \alpha\, e^{-\mymat{R}_1 \mymat{R}_2^\top}
  + (1-\alpha)\, e^{-\mymat{X}_1 \mymat{X}_2^\top}$ \label{algo-step:Cnode}
\State Perform $k$ iterations: $\mymat{C}_{\text{node}} \gets
  \bar{\mymat{A}}_1 \mymat{C}_{\text{node}} \ \bar{\mymat{A}}_2$ \label{algo-step:CnodePropagate}
\State $\mathbf{C}_{\text{anc}} \gets -\log\!\left(\varepsilon +\Call{OneHot}{\mathcal{L}}\right)$ \label{algo-step:Cancr}
 \State $\mymat{C}_1\gets(\mymat{R}_1\mymat{R}_1^{\top})\odot\bar{\mymat{A}}_1,\ \mymat{C}_2\gets(\mymat{R}_2\mymat{R}_2^{\top})\odot\bar{\mymat{A}}_2^T$  \label{algo-step:intra-cost}
\State Initialize $\mymat{S}$
\Repeat
  \Statex \textit{\quad // Phase 2: update $\mymat{S}$-dependent costs}
  \State $\mymat{C}_{\text{edge}} \gets
    \mymat{C}_1 \mymat{S} \mymat{C}_2^\top$  \label{algo-step:Cedge}
  \State $\mymat{C}_{\text{nbr}} \gets
    -\log\!\left(\bar{\mymat{A}}_1^\top \mymat{S} \bar{\mymat{A}}_2\right)$ \label{algo-step:Cnbr}
  \State $\mymat{C} \gets \mymat{C}_{\text{node}}
    + \lambda_e \mymat{C}_{\text{edge}}
    + \lambda_n \mymat{C}_{\text{nbr}}
    + \lambda_a \mymat{C}_{\text{anc}}$ \label{algo-step:Cfinal}
  \Statex \textit{\quad // Phase 3: solve for $\mymat{S}$}
    \RepeatTimes{$k$} \label{algo-step:sinkhorn}
     \State $\myvec{a}\gets\Call{SoftMin}{\mymat{C},\myvec{b}}$ 
      \State $\myvec{b}\gets\Call{SoftMin}{\mymat{C}^{\top},\myvec{a}}$ 
     \EndRepeat
     \State $\mymat{S}\gets\exp\!\big((\mathbf{1}\myvec{a}^{\top}+\myvec{b}\mathbf{1}^{\top}-\mymat{C})/\epsilon\big)$ \label{algo-step:updateS}
\Until{$\mymat{S}$ converges}
\Function{SoftMin}{$\mymat{C},\myvec{v}$}
  \State \Return $-\epsilon\log\!\big((\exp(-(\mymat{C}-\myvec{v}\mathbf{1}^{\top})/\epsilon))^{\top}\mathbf{1}\big)$ 
  % \Comment{fused shift $\oplus$ exp $\oplus$ sum}
\EndFunction
\end{algorithmic}
% \vspace{-10pt}
\end{algorithm}

Algorithm~\ref{alg:ot-align} takes two
networks with node features and a set of anchor alignments and
returns the alignment matrix $\mymat{S}$. After normalizing the
adjacency matrices into $\bar{\mymat{A}}_1, \bar{\mymat{A}}_2$
(Line~\ref{algo-step:norm}) and representing the anchors as one-hot
encodings $\mymat{R}_1 ^{(0)}\in \mathbb{R}^{n_1 \times |\mathcal{L}|}$ and
$\mymat{R}_2^{(0)} \in \mathbb{R}^{n_2 \times |\mathcal{L}|}$, the algorithm proceeds in three phases:

% Phase~1 computes the $\mymat{S}$-independent part of the cost matrix,
% Phase~2 computes the $\mymat{S}$-dependent components
% $\mymat{C}_{\text{edge}}$ and $\mymat{C}_{\text{nbr}}$, and Phase~3
% solves the OT problem.

\noindent\textbf{Phase 1: Computing $\mymat{C}_{\text{node}}$ (line 3-5).}
The node-consistency cost is built in three steps. First, repeated
multiplication 
% $\bar{\mymat{A}}_1 \mymat{R}_1$
$(1-\beta)\,\bar{\mathbf{A}}_i \mathbf{R}_i + \beta\,\mathbf{R}_i^{(0)}$
produces a personalized-PageRank-style 
% \qi{The current formulation of R1,R2 in Line 1-3 of Algorithm 1 seems more like a k-step graph propagation matrix rather than a personalized page-rank matrix in PARROT?}
positional encoding of each node with
respect to the anchor set; the $l$-th column of $\mymat{R}_i$ encodes
the position of every node relative to the $l$-th anchor
(Line~\ref{algo-step:Rk}). Second, Line~\ref{algo-step:Cnode} forms the initial
$\mymat{C}_{\text{node}}$ from these encodings via two matrix
multiplications. Finally, the cost is iteratively propagated across networks via the triple product
$\bar{\mymat{A}}_1 \mymat{C}_{\text{node}} \bar{\mymat{A}}_2$,
yielding the final $\mymat{C}_{\text{node}}$ that remains fixed
for the rest of the algorithm.
Computing $\mymat{C}_{\text{node}}$ thus reduces to six matrix
multiplications of distinct shapes: some involve sparse operands,
some are repeated, and some occur in tandem with others. 
We exploit these patterns in Section~\ref{sec:customSpMM}.

\noindent\textbf{Phase 1: Computing $\mymat{C}_{\text{anc}}$ (line~\ref{algo-step:Cancr}).}
The anchor-preference cost encodes prior supervision directly from the known correspondences. 
$\Call{OneHot}{\mathcal{L}}$ places a one at each anchor pair $(i,j)\in \mathcal{L}$ and zero elsewhere. 
% Shifting and applying negative $\log$, 
Each entry of $\mymat{C}_{\text{anc}}$ is $\approx 0$ on anchor pairs and $-\log(\varepsilon)$ otherwise, biasing transport toward known correspondences.

\noindent\textbf{Phase 2: Computing $\mymat{C}_{\text{edge}}$ (line 7 and 10).}
The edge-consistency cost is built in two steps. First, we form
intra-network dissimilarity matrices that capture how neighbors within
each network relate to one another with respect to the anchor set:
$\mymat{C}_1 = (\mymat{R}_1 \mymat{R}_1^\top) \odot \bar{\mymat{A}}_1$
and analogously for $\mymat{C}_2$ (Line~\ref{algo-step:intra-cost}). This triple product, called
\emph{sampled dense--dense matrix multiplication} (SDDMM), is less
expensive than a full dense product because only the entries
selected by $\bar{\mymat{A}}_1$ are computed. Since $\mymat{C}_1$
and $\mymat{C}_2$ do not depend on $\mymat{S}$, they are precomputed
in Phase~1. $\mymat{C}_{\text{edge}}$ is computed by transporting $\mymat{C}_1$ and $\mymat{C}_2$ across the networks via the triple product
$\mymat{C}_1 \mymat{S} \mymat{C}_2^\top$ (Line~\ref{algo-step:Cedge}).

\noindent\textbf{Phase 2: Computing $\mymat{C}_{\text{nbr}}$ (line~\ref{algo-step:Cnbr}).}
The neighborhood-consistency cost ensures that for any node pair
$(i,j)$, the alignment score $\mymat{S}_{ij}$ is consistent with the
alignment scores of $i$'s and $j$'s neighbors. It is computed by
propagating the current alignment across the two networks via the
triple product
$\mymat{C}_{\text{nbr}} = \bar{\mymat{A}}_1^\top \mymat{S} \bar{\mymat{A}}_2$.

\noindent\textbf{Phase 2: Computing $\mymat{C}$ (line 12).} After each cost term is computed, we add the matrices to obtain the final cost matrix for the current iteration. 
We also add the proximal term $\lambda_p \log(S)$, which couples the cost to the current alignment.

\noindent\textbf{Phase 3: Computing $\mymat{S}$ (line 13-17).}
We solve the entropic OT of Section~\ref{sec:ot-solver} by Sinkhorn (Line ~\ref{algo-step:sinkhorn}-15).
First, the dual vectors $\myvec{a}$ and $\myvec{b}$ are refined by k alternating SoftMin updates.
Each SoftMin is a log-sum-exp reduction over the cost matrix: $\myvec{a}$ reduces $\mymat{C}$ over its rows (Line~14), and $\myvec{b}$ reduces $\mymat{C}$ over its columns, equivalently a row reduction over $\mymat{C}^T$(Line~15).
Second, the alignment is updated from the dual vectors $\myvec{a}$ and $\myvec{b}$ (Line~\ref{algo-step:updateS}), where the outer products $\mathbf{1}\myvec{a}^{\top}$ and $\myvec{b}\mathbf{1}^{\top}$ broadcast the dual vectors into $n_1\times n_2$ matrices, combined with $\mymat{C}$, and exponentiated element-wise to produce $\mymat{S}$.

\subsection{Optimization opportunities}

The decomposition in Algorithm~\ref{alg:ot-align} exposes the
computation as a pattern of recurring dense and sparse linear-algebra
operations rather than as isolated steps, which lets us reason about
costs and bottlenecks directly. Assuming $n_1 \approx n_2 \approx n$~\cite{zeng2023parrot, yu2025joena},
Phase~1 incurs $\mathcal{O}(n^3)$ work across multiple matrix
multiplications; Phase~2 updates the $\mymat{S}$-dependent terms in
$\mathcal{O}(n^3)$ and produces a dense $\mathcal{O}(n^2)$ matrix in
each outer iteration; Phase~3 runs Sinkhorn at $\mathcal{O}(n^2)$ per
iteration. Memory traffic is a second bottleneck: the dense
$\mathcal{O}(n^2)$ cost and alignment matrices are streamed
repeatedly, rescaled row- and column-wise each Sinkhorn iteration,
and re-accessed during every cost update. Profiling PARROT, a representative
OT-based method, confirms that runtime is spread across all three
phases: fixed-cost computation $40\text{--}69\%$, cost update
$15\text{--}36\%$, and Sinkhorn $14\text{--}23\%$.

Because the three phases share a small set of recurring primitives,
accelerating those primitives accelerates the pipeline end-to-end.
But the primitives are diverse: the algorithm calls more than ten
distinct matrix multiplications differing in operand sparsity, shape,
and composition with adjacent operations. $\mymat{C}_{\text{node}}$
requires dense, sparse, and sparse-dense-sparse triple products;
$\mymat{C}_{\text{edge}}$ requires dense-dense-sparse (SDDMM) and
sparse-dense-sparse triple products; the OT solver requires
matrix--vector products interleaved with elementwise operations.
Off-the-shelf libraries such as Intel MKL deliver suboptimal performance across this mix. 
We identify three opportunities for closing the gap: (a) develop efficient sparse computations to reduce the runtime, (b) fuse a sequence of steps into a combined operation to reduce memory traffic, and (c) develop hardware-mapped parallelization for modern GPUs.

\subsection{Sparsity-aware Network Alignment}
\label{sec:sparsity}
Since real-world networks are sparse, representing graphs as sparse
matrices exposes asymptotically more efficient computations that
typically translate into faster runtimes in practice.
\textsc{FastAlign} represents the input graphs as sparse matrices and replaces every dense graph computation with a sparse kernel. 
% wherever an adjacency matrix is used in Algorithm~\ref{alg:ot-align}. 
%reducing the work from the number of node pairs to the number of edges.
The positional encoding (Line~3) and $\mymat{C}_{node}$ propagation (Line~5) multiply a sparse adjacency matrix with a dense matrix, which we cast as an SpMM.
Similarly, the sparse-dense-sparse triple product (Line~5) is factored as two successive SpMMs.
The same pattern recurs in the S-dependent costs $\mymat{C}_{edge}$ (Line~10) and $C_{nbr}$ (Line~11), which are also computed as two SpMMs.
We compute the intra-network dissimilarities (Line~7) using SDDMM that evaluates only the masked entries rather than materializing the full dense product.

\noindent\textbf{Computational Efficiency.}
These sparse substitutions change the asymptotic cost of the dominant operations. 
In the propagation and transport triple products (Lines 5, 10, 11), replacing GEMM with SpMM reduces computational complexity from $\mathcal{O}(n^3)$ to $\mathcal{O}(\nnz\cdot n)$, where $n$ is the number of nodes and \nnz~ is the number of nonzeros (edges) in the sparse matrix. For the intra-network dissimilarity (Line 7), each of the $\nnz$ masked entries is an inner product of length $|\mathcal{L}|$, reducing the cost from $\mathcal{O}(n^2\cdot |\mathcal{L}|)$ to $\mathcal{O}(\nnz\cdot |\mathcal{L}|)$.

% We exploit this sparsity while computing cnode , where structural and attribute information is propagated within each network and across networks using the normalized input graphs. 
% In cost update phase, we propagate computed alignment scores through the input graphs using sparse-dense-sparse matrix multiplication or sequence of SpMM operations. 
% In alignment propagation, replacing GEMM with SpMM reduces the cost from $\mathcal{O}(n^3))$ to $\mathcal{O}(nnz \cdot n))$ where $n$ is network size and $nnz$ is the number of nonzero values in adjacency matrix. 

% We also use sparse graphs structure as mask when calculating intra-network distance matrices in cedge, so distances are calculated only for current edges rather that all possible pairs of nodes.
% For computing intra-distance for neighboring nodes, SDDMM(Sampled Dense-Dense Matrix Multiplication) optimized to avoid full dense produce and computes only for entries guided by the sparse adjacency matrix, reducing the computational cost from $\mathcal{O}(n^2\cdot l)$ to $\mathcal{O}(nnz\cdot l)$ where $l$ is the number of anchor links.

\noindent\textbf{Memory Reduction.} \label{sec:precision} 
Utilizing sparse matrix representations scales down graph storage overhead from $\mathcal{O}(n^2)$ to $\mathcal{O}(\nnz)$.
We further observe that the alignment matrix is often sparse in practice because each node in one network typically has high alignment probability with only a small subset of nodes in the other network.~\cite{tang2023slotalign, yu2026avatar}
Thus, reducing the numerical precision
from double to single precision does not noticeably affect alignment accuracy. 
This also reduces the memory footprint and enables \textsc{FastAlign} to scale to larger graphs.

% In the sparse regime, the first two phases of \textsc{FastAlign} are dominated by
% sparse-dense matrix multiplication (SpMM), but with an unusual
% shape. Unlike the SpMM used in iterative solvers or GNNs where the
% dense operand is tall and skinny ($n \times d$, $d \ll n$), network alignment operates on an $n_1 \times n_2$ dense matrix, far wider than the settings considered in general-purpose SpMM kernels. 
% SDDMM and other products in the algorithm exhibit similarly atypical shapes. 
% We exploit this observations by developing customized operations.

\subsection{Custom SpMM for wide dense cost matrix} \label{sec:customSpMM}
Several of the most expensive steps in Algorithm~\ref{alg:ot-align} share a common matrix-product shape.
A dense cost matrix is multiplied on both sides by a sparse adjacency-structured matrix.
An example is the $\mymat{C}_{\text{node}}$ propagation $\bar{\mymat{A}}_1 \mymat{C}\,
\bar{\mymat{A}}_2^\top$ (Line~\ref{algo-step:CnodePropagate}), where $\mymat{C}$ is dense matrix of size $n_1 \times n_2$ and the outer matrices are sparse matrices $\bar{\mymat{A}}_1 \in \mathbb{R}^{n_1 \times n_1}$ and $\bar{\mymat{A}}_2 \in \mathbb{R}^{n_2 \times n_2}$.
The same sparse-dense-sparse structure recurs in the edge and neighborhood consistency terms $\mymat{C}_{\text{edge}} = \mymat{C}_1 \mymat{S} \mymat{C}_2^\top $ (Line~\ref{algo-step:Cedge}) and $\mymat{C}_{\text{nbr}} = \bar{\mymat{A}}_1^\top \mymat{S} \bar{\mymat{A}}_2$ (Line~\ref{algo-step:Cnbr}); there the outer factors $\mymat{C}_1, \mymat{C}_2$ are also sparse, with the same sparsity pattern as the adjacency matrix.
Computing each term requires two SpMMs that share the same
structure, a sparse $n \times n$ matrix multiplied by a dense matrix with $\Theta(n)$ columns.
We call this primitive \emph{SpMM on a wide dense matrix}.
Here \emph{wide} means \emph{not tall-skinny}: the number of columns in the dense matrix is comparable to or larger than its rows.
In our algorithm, this column count grows with the graph.
Even in the position-aware encoding (Line~\ref{algo-step:Rk}), which is another SpMM, $\mymat{R}_k$ has width $\approx 0.2\,n$, and it grows without bound as
the graphs grow.
This is the opposite of the regime that off-the-shelf
libraries such as Intel MKL is optimized for, where the dense matrix has a small, often \emph{fixed} column dimension $d$.
% It arises in the $\mymat{C}_{\text{node}}$ propagation (Line~\ref{algo-step:CnodePropagate}) and in the Edge and Neighborhood consistency terms $\mymat{C}_{\text{edge}}$ (Line~\ref{algo-step:Cedge}) and $\mymat{C}_{\text{nbr}}$ (Line~\ref{algo-step:Cnbr}); even the position-aware encoding (Line~\ref{algo-step:Rk}) which maps to SpMM , $\mymat{R}_k$ is comparatively wide, with width $\approx 0.2n_1 \approx 0.2n_2$.
% The natural way to compute each of these two SpMMs is to call an off-the-shelf libraries, which are optimized for a very different structure than our application, one where the dense operand is tall and skinny rather than wide. 
To see why that mismatch hurts us, we first review how a
generic SpMM is computed and why it is efficient in its intended regime.

% This primitive recurs throughout the algorithm, which requires two SpMM, yet general-purpose SpMM kernels serve it poorly. 
% \metacomment{blue}{todo}{Mention in batch rwr for large matrices d is not small. 20\% of ground truth. for 20k dataset d=l is 4000}

\noindent\textbf{Why generic SpMM underperforms.} 
A general SpMM ~\cite{yang2018design,qian2026sparsity,huang2020ge} computes $\mymat{M} = \mymat{A} \mymat{B}$ where $\mymat{A}$ is the sparse $n\times n$ matrix, and B and M are dense $n\times d$ matrices, where $d \ll n$. 
% In the first step, the position-aware cost propagation repeatedly applies $\mymat{C}_{cross} \gets \bar{\mat{G}}_1 \mymat{C}_{cross} \bar{\mat{G}}_2$. 
% In the second step, at every iteration we update the neighborhood- and structural-consistency regularization terms $\log\!\left(\bar{\mymat{G}}_1^{\top}\mymat{S}\bar{\mymat{G}}_2\right)$ and $\mymat{C}_1\mymat{S}\mymat{C}_2^{\top}$.
% In all of these computations, a wide dense matrix of size $n_1 \times n_2$ is multiplied on both sides by sparse matrices representing the two graphs' structure. 
% This primitive is costly and recurs extensively throughout the algorithm, yet it is poorly served by general-purpose sparse–dense (SpMM) kernels.
% General-purpose SpMM kernels are built for computing $\mymat{M} \gets \alpha \mymat{A} \mymat{B}$ where $\mymat{A}\in \mathbf{R}^{n_1\times n_1}$ is a sparse matrix and $\mymat{M}\in \mathbf{R}^{n_1\times d}$ and $\mymat{B}\in \mathbf{R}^{n_1\times d}$ are dense matrices.
% multiplying a sparse matrix of size  $n \times n$  with a tall and skinny dense matix of size $n \times d$ where $d \ll n$.
This kernel computes one output row at a time $\mymat{M}[i,:] = \sum_{k\,:\,A_{ik}\neq 0} A_{ik} \mymat{B}[k,:]$, 
% where for each row $i$ it captures nonzeros $A_{ik}$ and accumulates $\mymat{B}[k,:]$ into the output row $\mymat{M}[i,:]$. 
where $k$ runs only over the columns in which row $i$ of $\mymat{A}$ has a nonzero, $A_{ik}$ is that nonzero value, $\mymat{B}[k,:]$ is the $k$-th row of the dense input matrix, and $\mymat{M}[i,:]$ is the output row being accumulated.
These kernels are tuned for tall-skinny $\mymat{B}$ $(d\ll n)$ where a row $\mymat{B}[k,:]$ is short, stays cache-resident, and is reused across every nonzero that references column $k$ of $\mymat{A}$.
This setting is the most common application of SpMM when used in iterative solvers and graph neural networks.

% When $\mymat{B}$ is tall and skinny, each row $\mymat{B}[k,:]$ is small so an entire row stays cached and results in more cache reuse.
In \textsc{FastAlign}, $\mathbf{\mymat{B}}$ is wide, not tall-skinny. 
% For the cost products $d = n_2 \approx n_1$; even the position-aware encoding (Algorithm ~\ref{alg:ot-align}, Line~3), has $d\!\approx\!0.2\,n$.
A row $\mymat{B}[k,:]$ is now $\theta(n)$, so it may no longer stay cache-resident and cannot be reused across the nonzeros that reference column $k$ of $\mymat{A}$, which limits data reuse in \textsc{FastAlign}.
In the worst case, each nonzero from the sparse matrix fetches its corresponding dense row from memory, giving $\mathcal{O}(\nnz\cdot n)$ memory traffic.
Since SpMM is a memory-bound operation, the standard SpMM libraries (e.g., Intel MKL) hurt the performance of \textsc{FastAlign}.

 % With wide dense matrix $\mymat{B}$, this reuse is lost.
% Each row of $\mymat{B}$ can take up the whole cache line and the reuse distance exceeds the cache, and $\mymat{B}$ is effectively streamed once per nonzero.
% The SpMM kernel is memory-bounded since it has $\mathcal{O}(2 * NNZ * d$ memory access. 
% In our case the dense matrix is wide, $d = n_1 \approx n$, so this read grows to $\mathcal{O}(NNZ * n)$ and each accessed row is far too long to remain in cache. 
% The product becomes memory-bound and cache-unfriendly, which general-purpose kernels handle poorly.
% We therefore design a custom kernel that restores cache locality across the wide product.

\noindent\textbf{The fix: custom SpMM kernel.}
We design this kernel with a column blocking model to improve cache residency. 
We partition dense matrices $\mymat{B}$ and $\mymat{M}$ into $Q=\lceil n_2/b_c\rceil$ narrow blocks of $b_c$ contiguous columns and process one block at a time.
To compute block $q$ of output, whose columns start at $j_q=(q-1)b_c$, we fill it one output row at a time: $\mymat{M}[i, j_q:j_q+b_c]=  \sum_{k\,:\,\mymat{A}_{ik}\neq 0} \mymat{A}_{ik} \mymat{B}[k,j_q:j_q+b_c]$.
Since each memory access within each column block computation is limited to a narrow partition of dense matrix, the B partition needs to reside in cache, reduced to $n_1 \times b_c$ instead of $n_1 \times n_2$ that does not fit into L2/L3.
This will increase cache hits when consecutive rows share similar columns of nonzeros.

\begin{figure}
  \centering
  \includegraphics[width=0.95\linewidth]{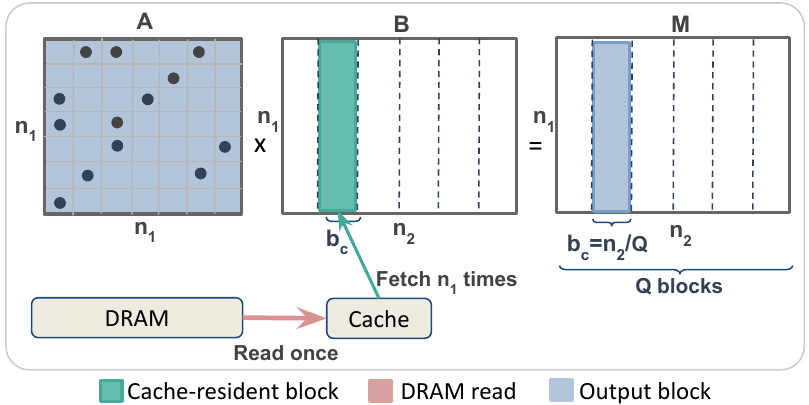}
  \caption{Column-blocked SpMM for one output block. To compute one $b_c$ block of $\mymat{M}$ (blue), the corresponding block of $\mymat{B}$ (green) is read from DRAM into cache once and reused across all rows of $\mymat{A}$.
% of $\mymat{B}$ for the blue block are loaded once into L2/L3 cache and reused across the block's rows, so $\mymat{B}$ is read from DRAM per block, not per nonzero.
}
  \label{fig:custom-spmm}
\end{figure}

Figure ~\ref{fig:custom-spmm} traces one block: 
To fill the $b_c$ output block $\mymat{M}[k,j_q:j_q+b_c]$, the kernel iterates over all rows of $\mymat{A}$ accumulating the $\mymat{A}_{ik} \mymat{B}[k,j_q:j_q+b_c]$ for each nonzero.
% for output row $i$, the kernel reads the nonzeros $\mymat{A}_{ik}$, gathers the corresponding slices $\mymat{B}[k,j_q:j_q+b_c]$ into the resident $n_1\times b_c$ panel, and accumulates them into $\mymat{M}[i,j_q:j_q+b_c]$.
% Because the panel stays in L2/L3, the next output row reuses those same slices from cache, and the only DRAM traffic for $\mymat{B}$ is the single load of the panel.
Since each memory access within a column-block computation is limited to a narrow partition of the dense matrix, the resident $\mymat{B}$ partition shrinks to $n_1\times b_c$, which fits into L2/L3, instead of the full $n_1\times n_2$ , which does not.
% \metacomment{purple}{ELAHEH}{Need to update this paragraph}
% \Ariful{Explain this with a figure. Do not undersell it. }

\noindent\textbf{Payoff.}
Blocking leaves the arithmetic and total access count unchanged at $\mathcal{O}(\nnz\cdot n)$.
It changes which level of memory is accessed.
By keeping each $\mymat{B}$ block cache-resident, it reduces the DRAM traffic for reading $\mymat{B}$, converting memory-bound stream into a cache-resident reuse pattern.
For a SpMM with the structure of our application, this improved cache reuse accelerates the SpMM computations throughout our model.

% \begin{figure}
%   \centering
%   \includegraphics[width=0.85\linewidth]{images/custom_spmm.png}
%   \caption{custom spmm}
%   \label{fig:custom-spmm}
% \end{figure}

% \textbf{Opportunity 2: Fusion of adjacent operations.}
% The cost update in Phase~2 and the Sinkhorn iterations in Phase~3
% consist of chains of elementwise operations and row- or column-wise
% reductions over dense $n_1 \times n_2$ matrices. Each step
% materializes a large dense intermediate that is repeatedly written
% and re-read, saturating memory bandwidth. Fusing each chain into a
% single pass eliminates these intermediates and the memory traffic
% they incur.
\subsection{Domain-specific Kernel Fusion}
\label{sec:fusion}
According to Algorithm ~\ref{alg:ot-align}, the cost matrix $\mymat{C}$ assembly (Line~12), and Sinkhorn dual vector update (Lines~14,15, Eq.~\ref{eq:sinkhorn-updates}) are written as compact algebraic expressions; however, a literal computation allocates and populates a $n_1 \times n_2$ intermediate matrix: each log, scale, and add streams a full dense intermediate to memory and back. 
Since these matrices are large, the memory traffic dominates the computation even though every individual operation is cheap.
We merge each algebraic chain into a single path over the output/input matrices.
We observed such chains that occur and fused them.

\noindent\textbf{Cost matrix $\mymat{C}$ computation.}
Cost computation (Line~12) combines four dense terms by scaling, log, and addition: $\mymat{C} \gets \mymat{C}_{\text{node}}
    + \lambda_e \mymat{C}_{\text{edge}}
    + \lambda_n \mymat{C}_{\text{nbr}}
    + \lambda_a \mymat{C}_{\text{anc}}$.
Naively computing this results in materializing at least 6 separate dense $n_1 \times n_2$ matrices. 
The fused kernel instead visits each entry $(i,j)$ once, loads the five contributing values, applies the scale/log/add, and writes $\mymat{C}_{ij}$ in one pass, without any intermediates. 

\noindent\textbf{Sinkhorn dual updates.}
Each dual update is a log-sum-exp reduction over the cost matrix (Eq.~\ref{eq:sinkhorn-updates}): the row reduction that updates a and the column reduction that updates b. 
Evaluated naively, each update shifts, exponentiates, and reduces the cost matrix, then applies a log and a final shift which materializes a full $n_1\times n_2$ intermediate matrix at each step. 
We instead design a single fused kernel that applies element-wise shift and exponential inside the reduction, in one streaming pass with a max-shift for stability, so the $n_1\times n_2$ intermediate is not materialized.
The same kernel serves both updates: applied to $\mymat{C}$ to refresh $\myvec{a}$ and to $\mymat{C}^T$ to refresh  $\myvec{b}$.

\subsection{GPU-aware Optimizations}

The computational reformulation of \textsc{FastAlign} naturally maps to GPU architectures, as the performance-critical matrix operations reduce to sparse and dense linear algebra kernels. 
Our GPU implementation builds on the sparsity-aware formulation (Section~\ref{sec:sparsity}) and domain-specific kernel fusion (Section~\ref{sec:fusion}) using cuSPARSE for sparse matrix operations and cuBLAS for dense linear algebra. 
Rather than using these libraries out of the box, we tune their performance by selecting memory layouts and SpMM algorithms to match the structure of each operation in the alignment pipeline.

\noindent\textbf{Minimizing latency.}
Unlike the CPU implementation, the GPU implementation must carefully minimize both device-side memory movement and host--device transfers to reduce latency.
Cost computation involves sparse--dense--sparse matrix products, which we break down into two SpMM calls with transposing the intermediate dense matrices.
As explicitly materializing transposed matrices is expensive, we use transpose views to avoid additional memory allocation.
To minimize host--device communication, all graphs, cost matrices, and alignment matrices remain resident in GPU memory, while intermediate buffers are reused throughout the alignment.

\noindent\textbf{GPU-specific kernel fusion.}
% The Sinkhorn dual updates assign one thread block per column and make two streaming passes --- one to reduce the column minimum using warp-level shuffles (\texttt{\_\_shfl\_down\_sync}), one to accumulate the shifted exponentials.
The Sinkhorn dual updates assign one thread block per column and execute two streaming passes: the first reduces the column minimum using warp-level shuffles (\texttt{\_\_shfl\_down\_sync}), while the second accumulates the shifted exponentials.
The same kernel serves the column reduction for updating $\mathbf{b}$ by operating on a transpose view of $\mathbf{C}$, produced by a tiled transpose kernel that stages $32\times32$ tiles in shared memory with $+1$ column padding to eliminate bank conflicts, achieving coalesced reads and writes in a single pass.

\noindent\textbf{Parallel streams and reusing SpMM plans.}
%FastAlign issues independent kernels  on separate graphs on parallel non-blocking CUDA streams when a single kernel leaves SM compute and bandwidth capacity underutilized.
To improve GPU utilization, \textsc{FastAlign} issues independent kernels for separate graphs across parallel, non-blocking CUDA streams.
For example, the row-normalization kernels of the two graph intra-distance matrices (Line~\ref{algo-step:intra-cost} of Algorithm~\ref{alg:ot-align}) run concurrently on separate streams rather than sequentially. %in a single fused kernel.
Furthermore, standard cuSPARSE SpMM calls incur per-invocation overhead for planning, handle setup, and buffer allocation. Because these calls repeat within a loop, \textsc{FastAlign} pre-computes the plan once and reuses it across iterations. This amortizes the initialization overhead, ensuring subsequent calls incur only raw execution costs.

\section{Evaluation}
\label{sec:evaluation}

We evaluate \textsc{FastAlign} to answer the following research questions:
\begin{itemize}
    \item[Q1] How accurate is \textsc{FastAlign} compared to SOTA network alignment algorithms?
    \item[Q2] How does \textsc{FastAlign} accelerate alignment on real networks?
    \item[Q3] How does \textsc{FastAlign} scale with increasing network size?
    \item[Q4] How much does each of our techniques contribute to the overall speedups achieved by \textsc{FastAlign}? 
\end{itemize}

\subsection{Experimental Setup}
For experiments, we use an AMD EPYC 7763 CPU with 64 cores and 512 GB of memory. 
For GPU experiments, we use an NVIDIA A100 GPU with 40 GB of memory and CUDA 12.8. 
All methods are evaluated in a semi-supervised setting, where 20\% of the ground-truth node pairs are used as prior knowledge, i.e., anchor links.
We report runtime and accuracy metrics averaged over five runs. 
Runtime measures the end-to-end alignment computation time after the input graphs are read, including all steps required to compute the final alignment matrix.
%We do not report separate convergence results because \textsc{FastAlign} preserves the original OT objective rather than changing the optimization problem. 
In our experiments, \textsc{FastAlign} exhibits convergence behavior similar to the OT baseline, PARROT.

\subsubsection{Metrics} 
We evaluate the effectiveness of \textsc{FastAlign} compared to the baselines using the following two accuracy metrics.
For each test node $x \in G_1$, let $x' \in G_2$ denote its ground-truth match. 
We rank all candidate nodes in $G_2$ by their alignment score with $x$.
A prediction is counted as a hit at $K$ if $x'$ appears among the top-$K$ ranked candidates.
For a test set with $n$ node pairs, $\mathrm{Hits@}K = \frac{\#\mathrm{hits@}K}{n}$.
Mean Reciprocal Rank (MRR) is measured as the average inverse rank of the ground-truth match: $\mathrm{MRR} = \frac{1}{n}\sum_{(x,x') \in \mathcal{S}_{\mathrm{test}}} \frac{1}{\operatorname{rank}(x,x')}$

\subsubsection{Baselines}
Following the benchmark study~\cite{yu2025planetalign}, we select six top-accuracy baselines that cover the three main categories of network alignment methods. 
FINAL~\cite{zhang2016final} is consistency-based; BRIGHT~\cite{yan2021bright} and NetTrans~\cite{zhang2020nettrans} are embedding-based; and JOENA~\cite{yu2025joena}, PARROT~\cite{zeng2023parrot}, and SLOTAlign~\cite{tang2023slotalign} are OT-based. 
PARROT is the closest baseline to \textsc{FastAlign}, since our framework preserves the same OT-based alignment formulation. 
We also evaluated NeXtAlign~\cite{zhang2021nextalign}, but it timed out on most datasets and is omitted.

\subsubsection{Datasets}
Table~\ref{tab:dataset-summary} summarizes the real-world datasets we used to evaluate \textsc{FastAlign} against the baselines. 
For scalability experiments, we generated synthetic graphs using the Erd\H{o}s--R\'enyi model with an average node degree of $10$~\cite{erdHos1960evolution}.

\begin{table}[t]
\centering
\caption{Dataset Characteristics}
\label{tab:dataset-summary}
\begin{tabular}{lrrrr}
\toprule
Networks & nodes & edges & attributes & sparsity (\%) \\
\midrule
Coral & 2,708 & 6,334 & 1,433 & 99.83\\
Coral2 & 2,708 & 4,542 & 1,433 & 99.88\\
\cmidrule(lr){1-5}
PPI1 & 3,480 & 117,429 & 50 & 98.06\\
PPI2 & 3,480 & 90,741 & 50 & 98.50\\
\cmidrule(lr){1-5}
ACM & 9,872 & 39,561 & 17 & 99.93\\
DBLP & 9,916 & 44,808 & 17 & 99.92\\
\cmidrule(lr){1-5}
GGI1 & 10,403 & 115,755 & 0 & 99.88\\
GGI2 & 10,403 & 89,448 & 0 & 99.91\\
\cmidrule(lr){1-5}
ArXiv1 & 18,722 & 217,921 & 0 & 99.6\\
ArXiv2 & 18,722 & 168,394 & 0 & 99.69\\
\cmidrule(lr){1-5}
DBP15K\_FR & 19,661 & 105,997 & 300 & 99.84\\
DBP15K\_EN & 19,993 & 115,722 & 300 & 99.83\\
\bottomrule
\end{tabular}
\end{table}

\begin{table*}[t]
\centering
\caption{Accuracy comparison with all baselines for real-world datasets on GPU (bold numbers denote the best accuracies).}
\begin{tabular}{l|cc|cc|cc|cc|cc|cc}
\hline
Dataset & \multicolumn{2}{c}{Cora} & \multicolumn{2}{c}{PPI} & \multicolumn{2}{c}{ACM-DBLP} & \multicolumn{2}{c}{GGI} & \multicolumn{2}{c}{ArXiv} & \multicolumn{2}{c}{DBP15K} \\
\cline{2-13}
Metrics & Hits@10 & MRR & Hits@10 & MRR & Hits@10 & MRR & Hits@10 & MRR & Hits@10 & MRR & Hits@10 & MRR \\
\hline
FINAL & 0.899 & 0.832 & 0.003 & 0.003 & 0.798 & 0.515 & 0.632 & 0.338 & 0.720 & 0.401 & 0.001 & 0.001 \\
\hline
BRIGHT & 0.674 & 0.470 & 0.629 & 0.511 & 0.774 & 0.504 & 0.588 & 0.456 & 0.695 & 0.492 & 0.617 & 0.404 \\
NetTrans & 0.716 & 0.486 & 0.791 & 0.634 & 0.800 & 0.526 & 0.677 & 0.499 & 0.785 & 0.551 & 0.642 & 0.408 \\
\hline
JOENA & \textbf{1.000} & 0.997 & 0.986 & 0.970 & 0.871 & 0.695 & 0.923 & 0.849 & 0.925 & 0.807 & \textbf{0.984} & \textbf{0.976} \\
PARROT & 0.965 & 0.964 & 0.998 & 0.994 & \textbf{0.947} & 0.776 & 0.933 & 0.855 & 0.918 & 0.796 & 0.970 & 0.886 \\
SLOTAlign & \textbf{1.000} & 0.997 & 0.980 & 0.965 & 0.874 & 0.745 & 0.803 & 0.740 & 0.724 & 0.607 & 0.848 & 0.774 \\
\hline
\textsc{FastAlign} & \textbf{1.000} & \textbf{0.997} & \textbf{0.999} & \textbf{0.996} & 0.943 & \textbf{0.840} & \textbf{0.941} & \textbf{0.879} & \textbf{0.929} & \textbf{0.811} & 0.916 & 0.799 \\
\hline
\end{tabular}
\label{tab:mrr-hits}
\end{table*}

\begin{figure*}
    % \vspace{-10pt}
    \centering
    \begin{minipage}{0.49\linewidth}
        \centering
        \includegraphics[width=\linewidth]{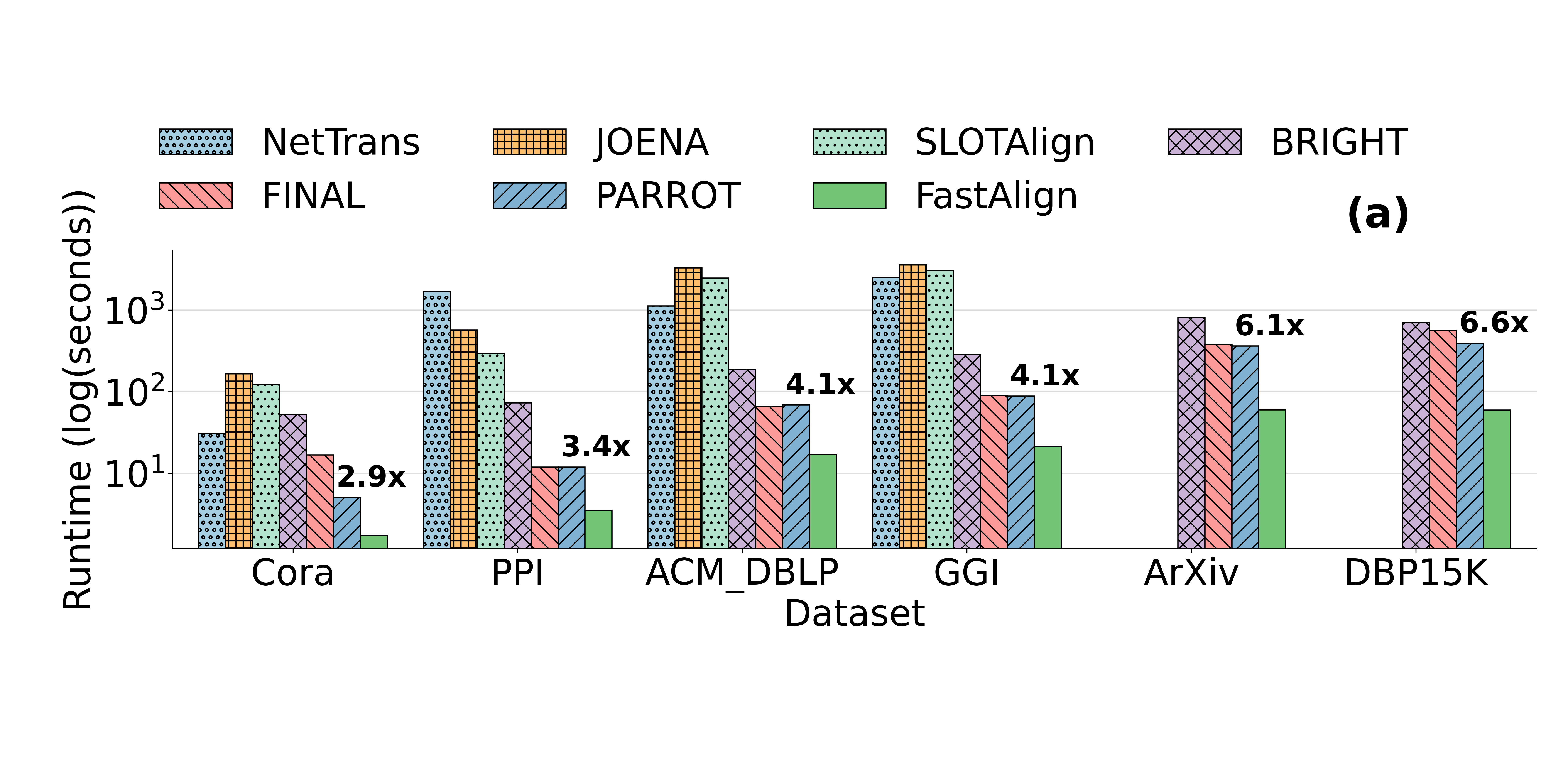}
            \label{fig:cpu_real_runtime}
    \end{minipage}
    \hfill
    \begin{minipage}{0.49\linewidth}
        \centering
        \includegraphics[width=\linewidth]{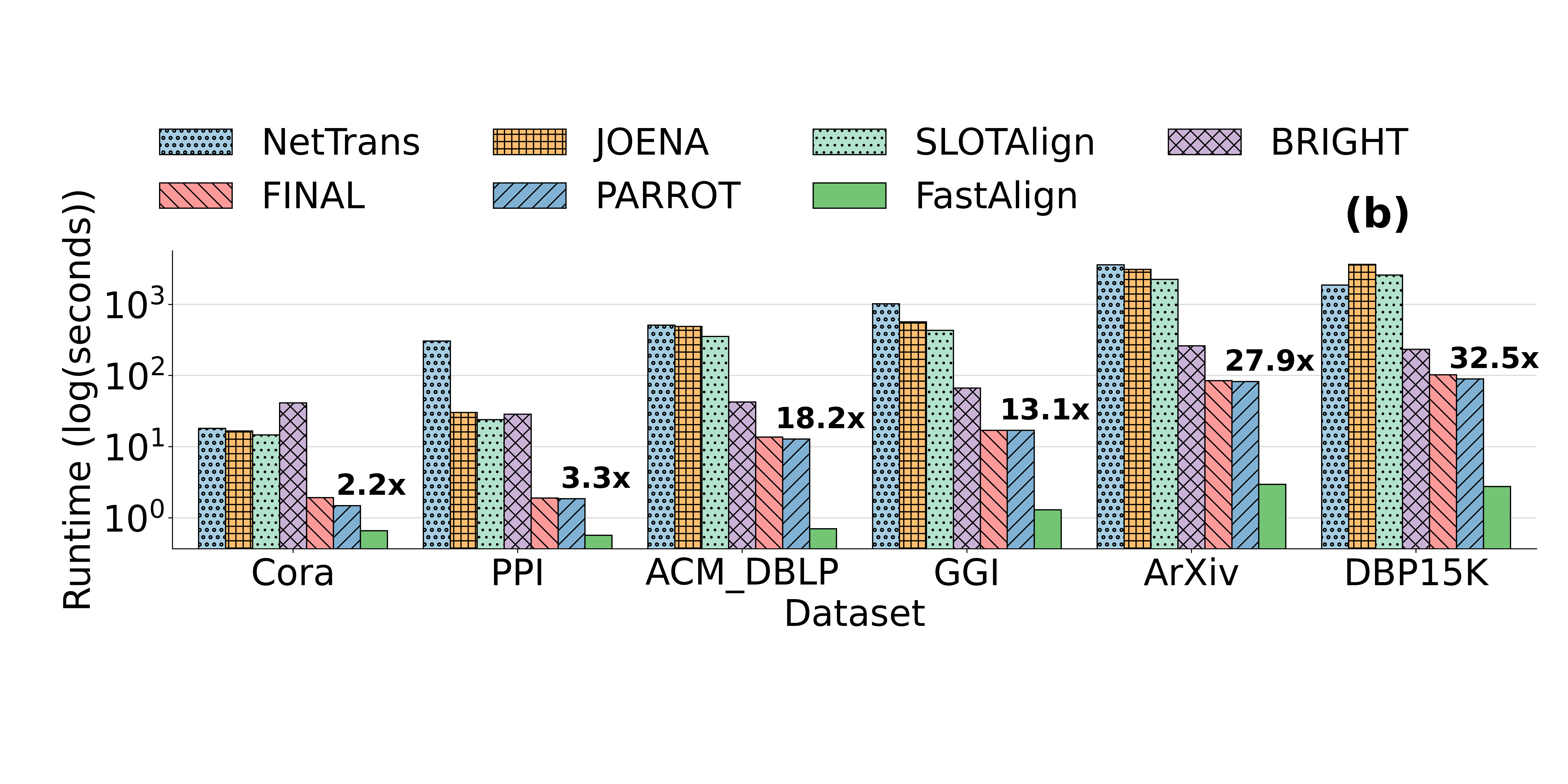}
            \label{fig:gpu_real_runtime}
    \end{minipage}
    \captionsetup{skip=-25pt}
    \caption{Runtime comparison of network alignment methods on real-world datasets on (a) CPU and (b) GPU architectures. Numbers on PARROT bar indicate speedups of \textsc{FastAlign} over PARROT.}
    \label{fig:real-runtime}
\end{figure*}

\subsection{Accuracy Results}
Table~\ref{tab:mrr-hits} compares the Hits@$10$ and MRR of \textsc{FastAlign} against all baselines on the real-world datasets. 
We report GPU results in this table and observe similar accuracy trends on CPU. 
CPU accuracy results are omitted because several baselines time out on larger datasets, as discussed in the next Subsection. 
\textsc{FastAlign} and other OT-based methods achieve substantially higher accuracy than consistency-based and embedding-based methods. 
Compared with state-of-the-art OT-based baselines, \textsc{FastAlign} achieves nearly identical or better accuracy on most datasets, showing that its optimizations do not degrade alignment quality and preserve original OT alignment objective. 
We attribute the slightly lower accuracy on DBP$15$K to floating-point roundoff inconsistencies, especially from fused kernels as they change the order of numerical operations.

\subsection{Performance Analysis}

% figure
% experiment
% trend
% reason
Figure~\ref{fig:real-runtime} compares the runtime of \textsc{FastAlign} against all baselines on real-world datasets using CPU and GPU implementations, respectively. 
On CPU, missing bars indicate that the corresponding baseline timed out on that dataset, whereas on GPU, all baselines completed. 
\textsc{FastAlign} achieves a $3.89\times$--$9.45\times$ speedup on CPU and a $2.24\times$--$32.54\times$ speedup on GPU compared to PARROT, the fastest baseline. 
Across all baselines, the speedup ranges from $3.89\times$ to $193.90\times$ on CPU and from $2.24\times$ to $1321.85\times$ on GPU, with the largest speedups achieved over slow baselines. 
%over the slowest embedding-based and OT-based baselines. 

% \begin{figure}[h]
%   \centering
%   \includegraphics[width=\linewidth]{images/cpu_real_runtime.png}
%   \caption{cpu real runtime}
%   \label{fig:cpu_real_runtime}
% \end{figure}

% \begin{figure}[h]
%   \centering
%   \includegraphics[width=\linewidth]{images/gpu_real_runtime.png}
%   \caption{gpu real runtime}
%   \label{fig:gpu_real_runtime}
% \end{figure}

These significantly high speedups demonstrate the effectiveness of \textsc{FastAlign}'s computational reinterpretation and optimizations discussed in Section~\ref{sec:design}.
In general, the speedups increase with dataset size. 
For datasets of similar size, the speedups follow the sparsity trend: higher sparsity leads to less data movement through memory and therefore larger speedups.
GPU speedups are higher than CPU speedups primarily because kernel fusion and GPU-specific optimizations have a stronger impact on GPU execution. 
In particular, kernel fusion improves data locality and reuse, reducing memory access latency and contributing to the superior performance of \textsc{FastAlign} on GPU.
% \metacomment{purple}{change this}{fusion has higher impact on gpu: fusing reduces letency, increases data reuse}
% For sparse input graphs, \textsc{FastAlign} replaces dense  GEMM operations with SpMM over the existing edges, i.e., the nonzero entries of the graph matrix. 
% This avoids unnecessary dense computation and memory movement, while kernel fusion further reduces redundant memory accesses.
% We discuss scaling and optimization performance breakdown in detail in the following sections.

\subsection{Scalability Analysis}
\begin{figure}[!t]
  \centering
  \vspace{-0.8em}
  \includegraphics[width=0.9\linewidth]{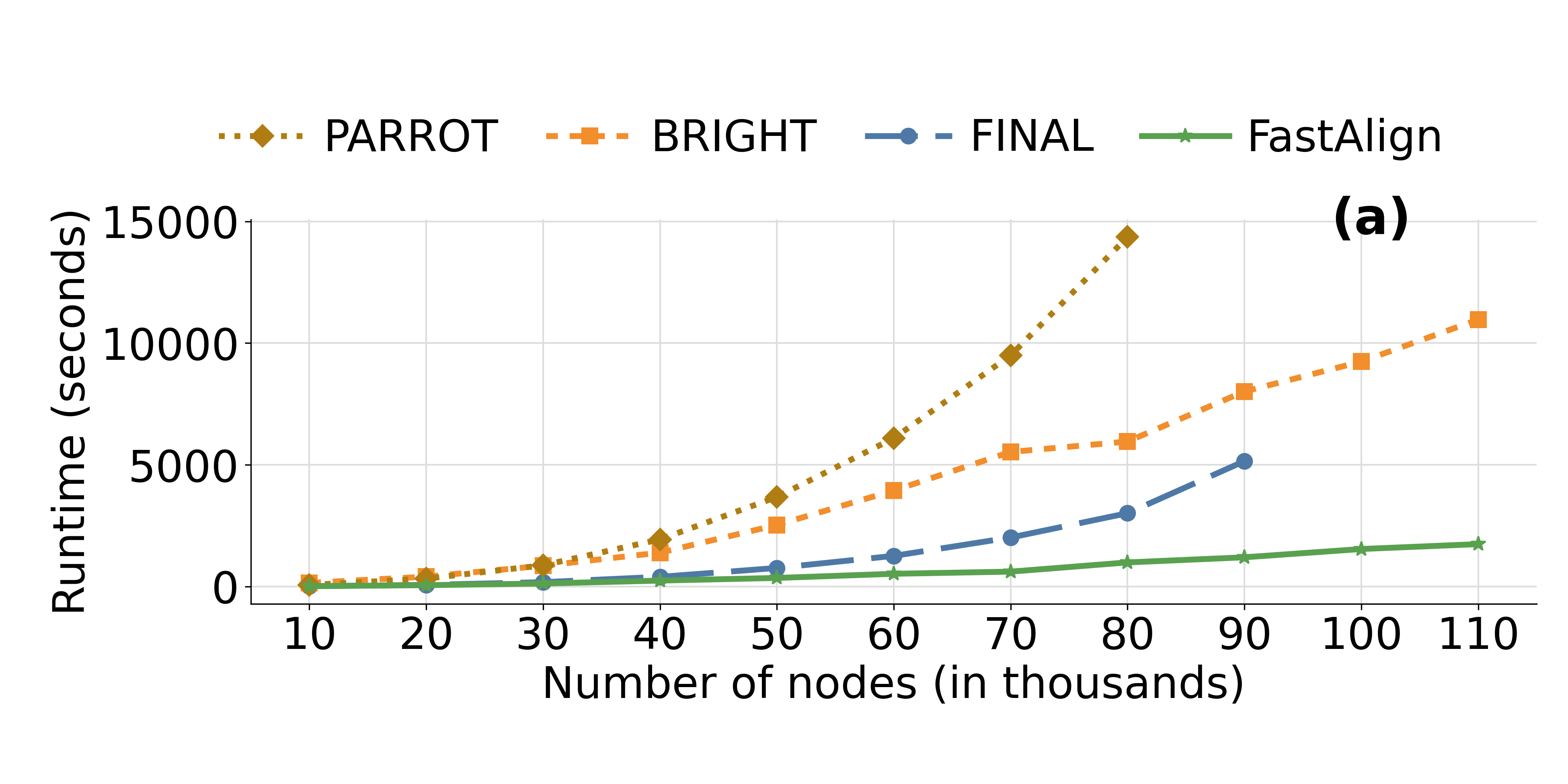}\\[-1.2em]
  \includegraphics[width=0.9\linewidth]{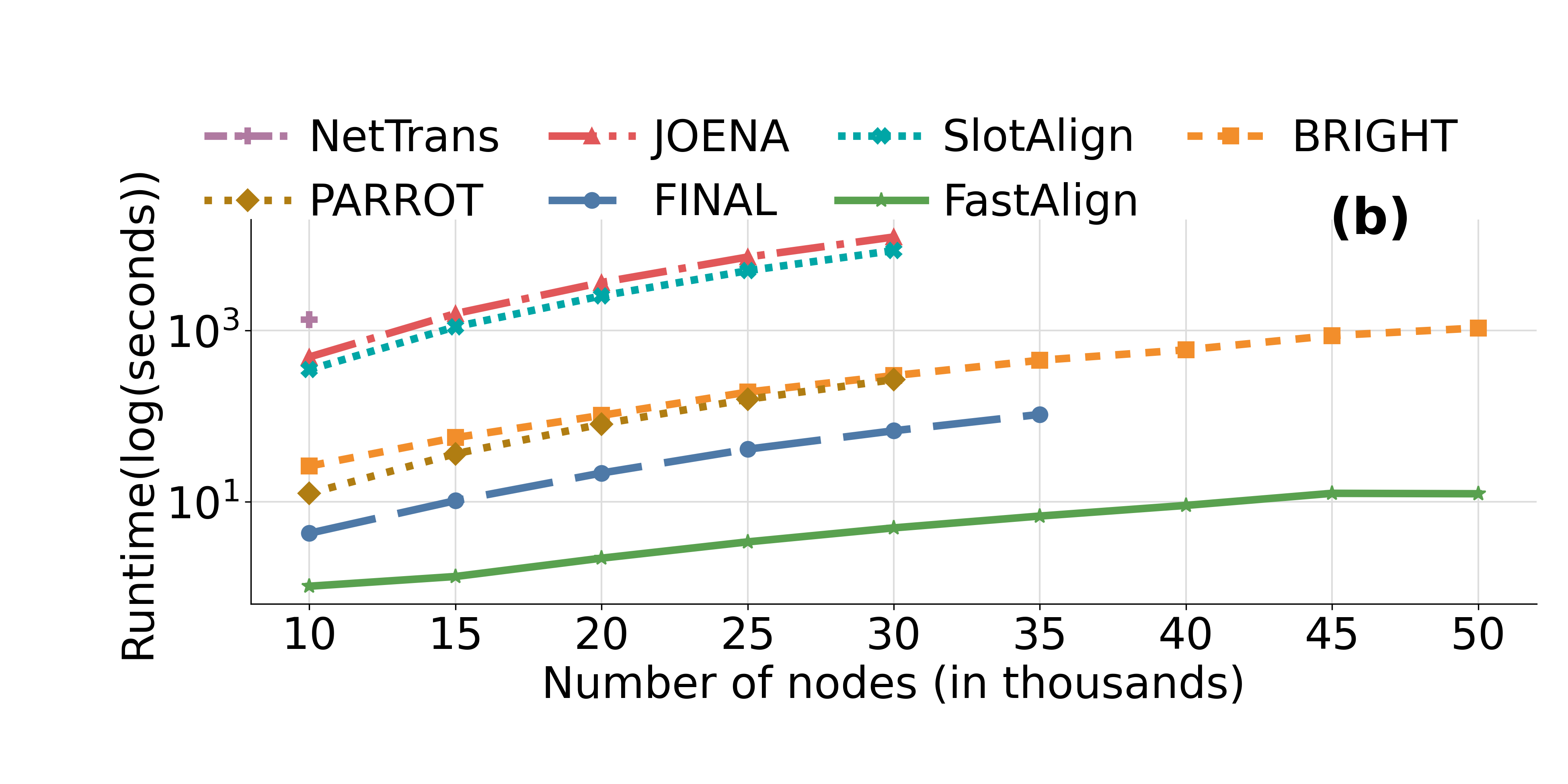}\\[0.5em]
   \captionsetup{skip=-5pt}
  \caption{Scalability comparison of network alignment methods on (a) CPU and (b) GPU architectures as the number of nodes in the synthetic graph increases.}
  \label{fig:er_runtime}
\end{figure}

Figure~\ref{fig:er_runtime}a shows the runtime scalability comparison of \textsc{FastAlign} for synthetic datasets against the baselines that did not time out on graphs with $20$K nodes. 
\textsc{FastAlign} scales more efficiently than all baselines, with speedups increasing as graph size grows --- from $0.82\times$ at 10K nodes to $9.41\times$ at 110K nodes over the fastest baseline.
The overall speedups are in the range $0.82 \times$ -- $14.49 \times$.

% \begin{figure*}[h]
%     \centering
%     \begin{minipage}{\linewidth}
%         \centering
%         \begin{subfigure}{0.49\linewidth}
%             \includegraphics[width=0.85\linewidth]{images/cpu_er_runtime.png}
%               \caption{cpu er runtime}
%               \label{fig:cpu_er_runtime}
%         \end{subfigure}
%         \hfill
%         \begin{subfigure}{0.49\linewidth}
%         \centering
%             \includegraphics[width=0.85\linewidth]{images/gpu_er_runtime.png}
%               \caption{gpu er runtime}
%               \label{fig:gpu_er_runtime}
%         \end{subfigure} 
%     \end{minipage}
%     \caption{synthetic graphs}
%     \label{fig:synthetic}
% \end{figure*}

\textsc{FastAlign} scales to graphs with up to $110$K nodes, while all baselines except \textsc{BRIGHT} run out of memory at smaller graph sizes.
The maximum problem size supported by \textsc{FastAlign} is determined by its peak memory usage during Sinkhorn optimization. 
The peak live state uses nine dense $(n_1 \times n_2)$ matrices and two sparse graph matrices. 
Using single precision (Section~\ref{sec:precision}), the dense matrices require $(4 \cdot 9 \cdot n_1 n_2 = 36n_1n_2)$ bytes, which bounds the implementation to approximately $110$K nodes on a CPU with $512$GB memory. 
As long as they fit in memory, \textsc{FastAlign} aligns the networks.

% \begin{figure}[h]
%   \centering
%   \begin{minipage}{\linewidth}
%     \centering
%     \includegraphics[width=0.9\linewidth]{images/cpu_er_runtime_inside.png}
%     \label{fig:cpu_er_runtime}
%   \end{minipage}
%   % \vspace{-1pt}
%   \begin{minipage}{\linewidth}
%     \centering
%     \includegraphics[width=0.9\linewidth]{images/gpu_er_runtime_inside.png}
%     \label{fig:gpu_er_runtime}
%   \end{minipage}
%   % \captionsetup{skip=4pt}
%   \caption{Scalability comparison of network alignment methods on (a) CPU and (b) GPU architectures as the number of nodes in the synthetic graph increases.}
%   \label{fig:er_runtime}
% \end{figure}

% \begin{figure}[h]
%   \centering
%   \includegraphics[width=0.78\linewidth]{images/cpu_er_runtime.png}
%   \caption{cpu er runtime}
%   \label{fig:cpu_er_runtime}
% \end{figure}

% \begin{figure}
%   \centering
%   \includegraphics[width=0.8\linewidth]{images/gpu_er_runtime.png}
%   \caption{gpu er runtime}
%   \label{fig:gpu_er_runtime}
% \end{figure}

\begin{figure*}
    \centering
    \begin{minipage}{0.49\linewidth}
        \centering
        \includegraphics[width=\linewidth]{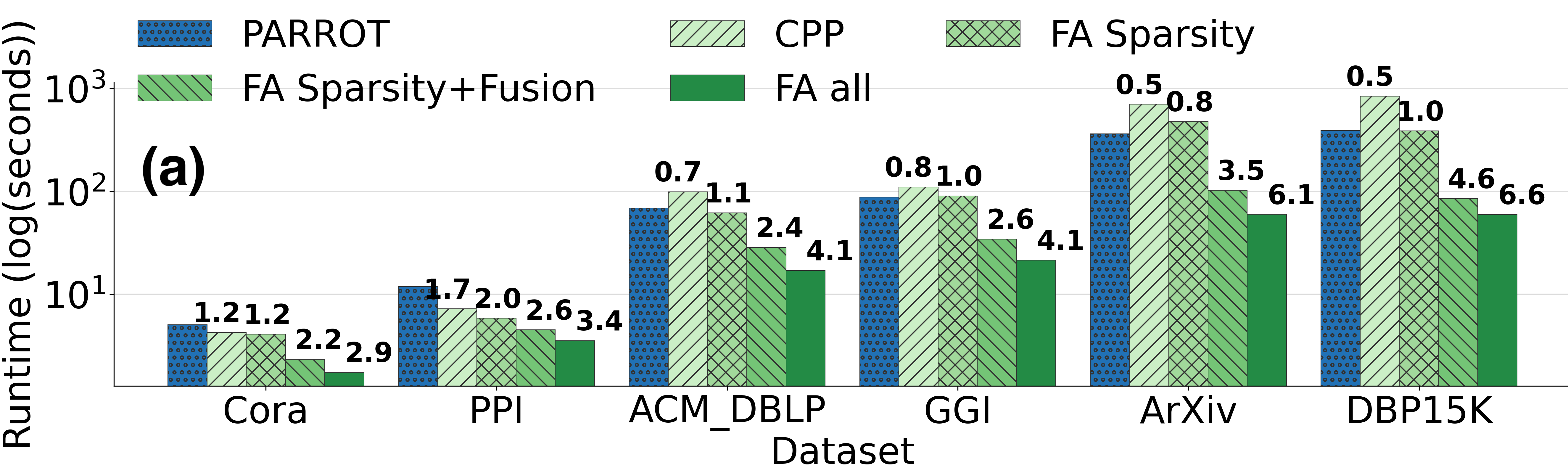}
        \label{fig:cpu_ablation}
    \end{minipage}
    \hfill
    \begin{minipage}{0.49\linewidth}
        \centering
        \includegraphics[width=\linewidth]{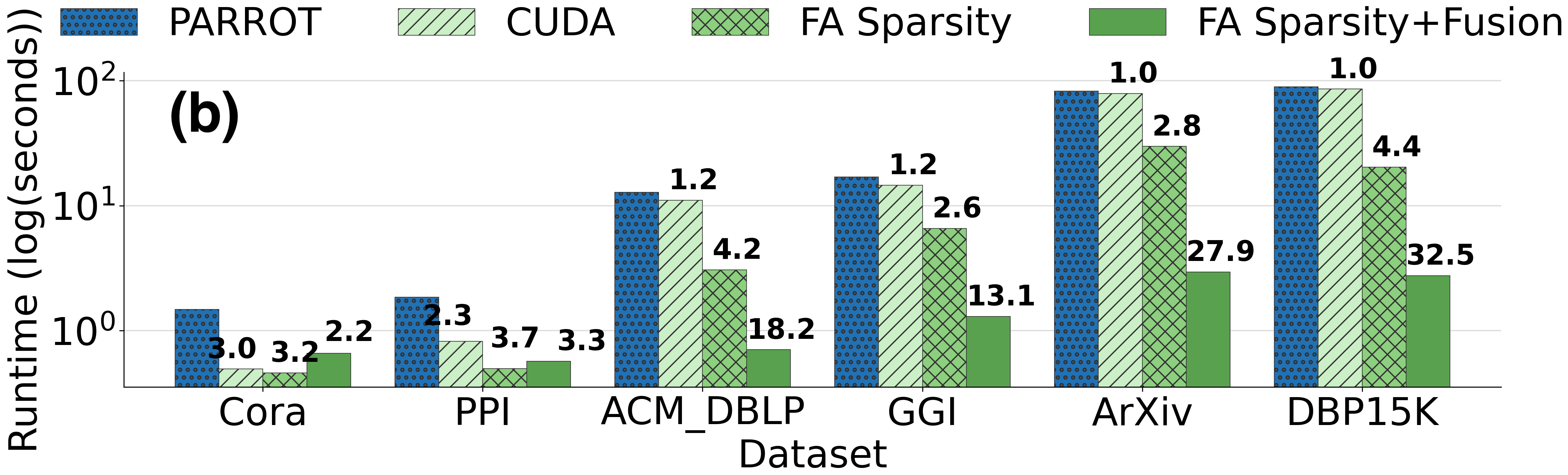}
        \label{fig:gpu_ablation}
    \end{minipage}
    \captionsetup{skip=-6pt}
    \caption{Performance contribution of \textsc{FastAlign}'s optimizations to its overall runtime on (a) CPU and (b) GPU architectures for real-world networks. Numbers on bars indicate speedups relative to \textsc{PARROT}.}
    \label{fig:ablation}
\end{figure*}

Figure~\ref{fig:er_runtime}b presents the scalability comparison on GPU using an NVIDIA A100 (80GB variant). 
\textsc{FastAlign} and \textsc{BRIGHT} scale to 50K-node graphs, while \textsc{FINAL} supports up to 35K nodes, \textsc{PARROT}, \textsc{JOENA}, and \textsc{SLOTAlign} up to 30K, and \textsc{NetTrans} only up to 10K.
Although the y-axis is shown in log scale to accommodate the wide range of runtimes, the scalability trend on GPU mirrors the CPU results: \textsc{FastAlign}'s speedups increase with graph size. 
From 10K to 50K nodes, it achieves speedups in the range of $4.16\times$--$86.88\times$ over \textsc{BRIGHT}, the fastest baseline.
Across all baselines and graph sizes, the speedups are in the range of $4.16\times$--$2525.88 \times$.

\subsection{Ablation studies}
Figure~\ref{fig:ablation} shows the runtime contribution of each \textsc{FastAlign} optimization: sparsity-aware computation, kernel fusion, and custom SpMM.
The number on each bar reports the speedup of that artifact over PARROT.
To isolate algorithmic gains from platform effects, we include a straightforward reimplementation of the baseline, PARROT in CPP and CUDA as an intermediate baseline. 
Both CPP and CUDA baselines are only marginally faster than the original PyTorch implementation of PARROT, with the speedups diminishing as graph size grows, confirming that naive reimplementations with a platform change yield no meaningful performance benefit.
On smaller networks, we observe a modest speedup from framework conversion alone.
However, on larger datasets with at least $10$K nodes, the overhead of PyTorch over CPP/CUDA becomes negligible and the C++/CUDA baselines are no longer faster, confirming that \textsc{FastAlign}'s speedups are driven by its algorithmic optimizations rather than the platform change.

On CPU, sparsity alone improves runtime by $0.8\times$--$2.6\times$, and adding fusion compounds these gains to $2.2\times$--$4.6\times$, with custom SpMM further increasing speedups to $2.9\times$--$6.6\times$. 
On GPU, sparsity contributes $2.6\times$--$4.4\times$ to the speedup, and adding fusion brings the overall speedup to $2.2\times$--$32.5\times$.
Across both CPU and GPU, all optimizations contribute to performance, with fusion consistently compounding the benefits of sparsity.

% \begin{figure}[h]
%   \centering  \includegraphics[width=0.9\linewidth]{images/gpu_ablation.png}
%   \caption{Ablation study on CPU}
%   \label{fig:cpu_ablation}
% \end{figure}

% \begin{figure}[h]
%   \centering
% \includegraphics[width=0.9\linewidth]{images/gpu_ablation.png}
%   \caption{Ablation study on GPU}
%   \label{fig:gpu_ablation}
% \end{figure}

% GPU speedups are higher than CPU speedups because the matrix operations in network alignment map well to GPU architectures and are generally well optimized on GPUs, especially through cuBLAS and cuSPARSE, while the remaining kernels are tiled and hand-optimized.

% While kernel fusion reduces the redundant mempry accesses, 

\section{Related Work}
\label{sec:related_work}

In this section, we review the related work, which can be categorized into two groups: network alignment and optimal transport.

\subsubsection{Network Alignment}
Research on network alignment has largely developed along two directions: consistency-based and embedding-based approaches. 
The first direction is based on the idea that aligned nodes should have structurally and semantically similar positions, so that the connectivity and attributes of a node's neighbors carry over from one network to the other. 
IsoRank ~\cite{singh2008global} propagates pairwise similarities on the product graph, FINAL ~\cite{zhang2016final} extends this with attribute and edge consistency for attributed graphs, and
% , and MOANA ~\ref{} performs alignment hierarchically across coarsened views to improve efficiency. 
BIG-ALIGN ~\cite{koutra2013big} recovers a near-permutation between adjacency matrices via a Frobenius objective.
Computationally, these methods iterate over an $\mathcal{O}(n^2)$ product or similarity space that dominates cost at scale, motivating coarsening-based designs like MOANA ~\cite{zhang2019multilevel}.
In terms of accuracy, however, they capture limited global graph geometry, and the consistency premise weakens when the two networks differ substantially ~\cite{zhang2021nextalign}.

The second learns node embeddings that encode each network's topology while pulling anchor pairs together:
IONE~\cite{liu2016aligning} models directional follower/followee context, REGAL~\cite{heimann2018regal} factorizes a cross-network similarity matrix, CrossMNA~\cite{chu2019cross} fuses intra-/inter-/global features, and NetTrans~\cite{zhang2020nettrans}, BRIGHT~\cite{yan2021bright}, and NeXtAlign~\cite{zhang2021nextalign} further improve robustness through learned transformations, restart-based positional features, and consistency-aware sampling, respectively. 
Computationally, the cost shifts to training: repeated gradient updates over deep encoders, often with pair sampling, are expensive and lack convergence guarantees.
In accuracy, they model cross-network relations only indirectly through anchor paths and are sensitive to sampling noise and embedding-space distortion~\cite{yu2025joena,zhang2021nextalign}.

\subsubsection{Optimal Transport}
Optimal transport (OT) has recently been applied to network alignment, representing each graph as a distribution over its nodes and seeking the minimum-cost transport between them~\cite{chen2020graph}. 
Methods differ mainly in how the transport cost is defined.
Early designs transport spectral node embeddings (EMD~\cite{nikolentzos2017matching}) or represent graphs through filtered graph signals and Gaussian graph kernels (FGOT~\cite{maretic2022fgot}, GOT~\cite{petric2019got}), while later costs jointly encode node and edge correspondence via the Wasserstein and Gromov–Wasserstein (GW) distances~\cite{xu2019scalable,chen2020graph}.
A prominent line aligns graphs through Gromov–Wasserstein structural discrepancy (S-GWL~\cite{xu2019scalable}, GraphOTC~\cite{o2021graph}), while GOAT~\cite{saad2021graph} accelerates it via the Sinkhorn algorithm.
More recent work learns the cost instead of fixing it: SLOTAlign~\cite{tang2023slotalign} encodes a GW objective with a parameter-free GNN, CombAlign~\cite{chen2025combalign} ensembles embedding- and OT-based alignments, and JOENA~\cite{yu2025joena} jointly learns embeddings and transport. 
PARROT~\cite{zeng2023parrot} builds random-walk topology and consistency regularization into the OT cost and solves it with a constrained proximal point method.

While OT-based methods achieve higher accuracy, they remain the most computationally demanding category.
Their cost stems from repeatedly updating dense $\mathcal{O}(n^2)$ matrices, and while proximal solvers~\cite{yu2025joena} reduce iteration counts, graph sparsity and sparse-linear-algebra primitives remain underexploited.
As a result, even the most scalable OT aligners are confined to moderate-sized graphs, a gap our work targets by retaining the accuracy of regularized OT while improving runtime and scalability through a computation-optimized algorithm.

\section{Conclusion}
\label{sec:conclusion}
In this paper, we scale OT-based network alignment to large graphs while preserving its accuracy.
Instead of designing a new alignment model, we present \textsc{FastAlign}, which keeps the original OT formulation and treats its computation as a small set of recurring sparse-dense operations.
\textsc{FastAlign} exploits graph sparsity to skip work over nonexistent edges, fuses memory-bound kernels to cut redundant memory traffic, adds a custom SpMM kernel for the wide dense matrices in OT alignment, and uses GPU-aware optimizations to reuse data on device.
Experiments on real and synthetic datasets show that \textsc{FastAlign} matches the accuracy of SOTA OT-based methods while running up to $9.45\times$ faster on CPU and up to $32.54\times$ faster on GPU than the strongest OT baseline.

% \section*{Acknowledgments}

% \begin{acks}
% We are thankful to .
% \end{acks}

% \balance
\bibliographystyle{ACM-Reference-Format}
\bibliography{references}

\end{document}